\documentclass[10pt]{article}
\usepackage{blindtext}
\usepackage[a4paper, total={6in, 8in}]{geometry}
\usepackage[numbers]{natbib}
\usepackage{lastpage}
\usepackage[utf8]{inputenc} 
\usepackage[T1]{fontenc}    
\usepackage{hyperref}       
\usepackage{url}            
\usepackage{booktabs}       
\usepackage{amsfonts}       
\usepackage{nicefrac}       
\usepackage{microtype}      
\usepackage{xcolor}         
\usepackage{color,soul}
\usepackage{threeparttable}
\usepackage{array,booktabs} 
\usepackage{float}
\usepackage{adjustbox}
\usepackage{longtable}
\usepackage{microtype}
\usepackage{graphicx}
\usepackage[ruled,vlined,nofillcomment]{algorithm2e}
\usepackage{mathtools,amsmath,amsthm,amssymb,amsfonts,bm}
\usepackage{colortbl}
\usepackage{multirow}
\usepackage{tikz}
\usepackage{subcaption}
\usepackage{pgfplots}
\usepackage{enumitem}
\pgfplotsset{compat=newest}
\pgfplotsset{plot coordinates/math parser=false}
\usetikzlibrary{shapes,arrows,fit,calc,positioning,matrix}
\usetikzlibrary{shapes.geometric}
\usetikzlibrary{spy}
\tikzset{internal/.style={draw, ellipse, x radius=3cm, y radius=3cm, thick, text centered}}
\tikzset{subtree/.style={isosceles triangle, isosceles triangle apex angle=80, draw, shape border rotate=90, thick, text centered,minimum height=1cm}}
\tikzset{leaf/.style={draw, rectangle, thick, text centered, minimum height=0.5cm, minimum width=1cm}}
\tikzset{line/.style={draw, thick, -latex'}}
\tikzset{state/.style={draw, circle, fill=white,draw=black,thick,text centered}}
\tikzset{cell/.style={rectangle, thick, draw}}

\hypersetup{
    colorlinks=true,       
    linkcolor=black,          
    citecolor=black,        
    urlcolor=blue           
}

\newcommand{\T}{\mathbb{T}} 
\newcommand{\Lf}{\mathcal{L}} 
\newcommand{\I}{\mathcal{I}} 
\newcommand{\X}{\mathbf{X}} 
\newcommand{\Y}{\mathbf{Y}} 

\begin{document}

\title{RJHMC-Tree for Exploration of the Bayesian Decision Tree Posterior}
\author{ Jodie A.~Cochrane \\\small \texttt{Jodie.Cochrane@newcastle.edu.au} 
   \and
  Adrian Wills \\ \small\texttt{Adrian.Wills@newcastle.edu.au } 
   \and
   Sarah J.~Johnson \\ \small\texttt{Sarah.Johnson@newcastle.edu.au} 
}

\date{}

\maketitle

\begin{abstract}

Decision trees have found widespread application within the machine learning community due to their flexibility and interpretability.
This paper is directed towards learning decision trees from data using a Bayesian approach, which is challenging due to the potentially enormous parameter space required to span all tree models.
Several approaches have been proposed to combat this challenge, with one of the more successful being Markov chain Monte Carlo (MCMC) methods.
The efficacy and efficiency of MCMC methods fundamentally rely on the quality of the so-called proposals, which is the focus of this paper.
In particular, this paper investigates using a Hamiltonian Monte Carlo (HMC) approach to explore the posterior of Bayesian decision trees more efficiently by exploiting the geometry of the likelihood within a global update scheme.
Two implementations of the novel algorithm are developed and compared to existing methods by testing against standard datasets in the machine learning and Bayesian decision tree literature. HMC-based methods are shown to perform favourably with respect to predictive test accuracy, acceptance rate, and tree complexity.

\end{abstract}

\section{Introduction}
\label{sec:introduction}
Decision trees are ubiquitous in the machine learning community due to their model interpretability and flexibility to describe relations in the data. Many areas, for example in clinical settings, prioritise such models due to the straightforward connection between the final model and the original dataset. A standard decision tree defines a set of ordered splits that are applied sequentially to the data. The traditional approach to learning a decision tree is via a greedy one-step ahead heuristic. One of the most popular of such methods is known as CART \citep{breiman1984classification} which gives the output predicted by a tree model as a point-estimate.

Recently, research has focused on probabilistic approaches to learning these models \citep{ghahramani2015probabilistic}, since they offer quantification of model uncertainty.
Within this broad class of approaches, Bayesian inference has proven effective but comes with added complexity in computing the required distributions. Markov chain Monte Carlo (MCMC) methods are commonly used in Bayesian inference to approximate the desired probability distribution using the law of large numbers \citep{tierney1994markov}, whose convergence rate depends on how uncorrelated the samples are. The performance of these methods is therefore directly related to the quality of the samples in the chain - where quality relates to both how uncorrelated the chain is and that the samples are in areas of high likelihood. The effectiveness of MCMC methods is critically linked to the quality of samples from the Markov chain, and while attempts have been made, providing quality samples remains an open challenge and is the primary focus of this paper.

The central difficulty is that the tree structure must be explored along with the other tree parameters. This is challenging for MCMC methods because the problem dimension changes with the tree structure, a quality referred to as transdimensional. Exploring the posterior of Bayesian decision trees was first investigated in \citet{buntine1992learning}, which showed that using a Bayesian averaging and smoothing approach often produced more accurate predictions than heuristic approaches. Shortly thereafter, a stochastic sampling method based on MCMC techniques was independently proposed by \citet{chipman1998bayesian} and \citet{denison1998bayesian}, both of which developed a set of tree-inspired moves (i.e. grow, prune, change and swap) to transition from one tree to the next, with each tree constituting a new sample in the chain.
This again saw an improvement in the accuracy of the trees that were visited, including some with better predictive ability than standard non-Bayesian methods. 

Further developments in the area witnessed additions and improvements made to the original set of tree-inspired moves within the MCMC framework. In particular, \citet{wu2007bayesian} introduced a new type of move, referred to as the `radical restructure'. 
\citet{gramacy2008bayesian} used binary search tree theory to propose an improved version of the original swap move.  
\citet{pratola2016efficient} then generalised this improved swap move to a broader range of scenarios, in addition to improving on the original change move.

Stochastic exploration has also been tackled using a different method for sampling candidate decision trees called Sequential Monte Carlo (SMC). In contrast to MCMC methods, where a single tree is altered at every iteration, SMC works by sequentially changing a set of particles at every iteration and weighting based on the likelihood, where in this case each of these particles represents a decision tree. This solution was first investigated by \citet{taddy2011dynamic} in the context of an increasing number of data observations, where the decision tree was changed locally based on the location of the new datapoint. \citet{lakshminarayanan2013top} applied SMC to construct a Bayesian decision tree by instead using the entire dataset and proposing locally optimal moves. 

The main issue with the above methods is that a local random-walk mechanism is often used to propose a new tree, which does not consider how likely the proposed tree is with respect to the data. The only exception to this is the optimal proposal presented in \citet{lakshminarayanan2013top} whereby a local likelihood is used to propose a new split value. However, this still remains a local-based proposal to the current tree. Using a localised random-walk scheme as the MCMC proposal mechanism can result in poor acceptance rates, as the random perturbation in the tree is often not likely with respect to the data. For SMC, this issue presents itself as particle degeneracy, where the likelihood-based weighting system may select only a few trees considered likely at each iteration. The consequence of proposing unlikely samples is that the acceptance rate is low, causing the chain to suffer from slow convergence and poor mixing, and subsequently limiting the exploration of the posterior distribution.

The method proposed in this paper attempts to improve acceptance rates and predictive accuracy by incorporating Hamiltonian Monte Carlo (HMC) sampling into the tree proposal scheme. HMC is an MCMC-based method that is known to be efficient at sampling as it uses the geometry of the likelihood to propose the next sample \citep{betancourt2017conceptual}. These trajectories are defined using the derivative of the so-called `potential energy function' which in this case is taken to be the negative of the log-posterior of the decision tree parameters. This enables the proposed sample for continuous parameters to be jointly updated and informed by the geometry of the posterior distribution, improving the efficiency of exploration. 

However, it is not straightforward to apply HMC to decision trees as the hard splits of the decision process result in a piecewise constant (and therefore not everywhere-differentiable) likelihood. To overcome this pathology, we introduce two new parameterisations such that HMC can be applied: first, additional adaptation-phase hyperparameters, motivated by the work of \citet{linero2018bayesian}, which softens the hard split constraint, and second,  an input selection method that softens the splitting index parameter. As will be shown in this paper, these methodologies appear to help mitigate the low acceptance rate that is common in current posterior exploration methods and, in most cases, achieve better testing accuracy simultaneously with a simpler tree structure.

This paper is organised as follows: Section 2 establishes the decision tree model, associated notation and problem definition. Section 3 then presents the novel RJHMC-Tree algorithm, with details on the two implementations, overall proposal scheme and acceptance probabilities. The algorithm is then tested against synthetic and real-world datasets and compared to existing approaches in Section 4. Section 5 summarises the novel approach with a discussion and concluding remarks.

\section{Decision tree model and problem definition}
\label{sec:model}
The dataset $\mathcal{D} = \{ (\mathbf{x}_i, \mathbf{y}_i )\}_{i=1}^N$ represents the set of $N$ observations of inputs $\mathbf{x}_i \in \mathcal{X}$ and corresponding outputs $\mathbf{y}_i \in \mathcal{Y}$. Each data point $(\mathbf{x}_i, \mathbf{y}_i )$ represents a vector of $n_x$ inputs $\mathbf{x}_i = [ \begin{matrix}x_{i,1} & x_{i,2} & \cdots & x_{i,n_x} \end{matrix} ]^T$ and a set of $n_y$ corresponding outputs $\mathbf{y}_i = [ \begin{matrix} y_{i,1} & y_{i,2} & \cdots & y_{i,n_y} \end{matrix} ]^T$. We will consider both classification and regression problems, i.e., $\mathcal{Y} = \{1, 2, \dots, M  \}$ where $M$ denotes the number of possible discrete outputs or $\mathcal{Y} = \mathbb{R}^{n_y}$ where $n_y$ denotes the dimension of the output space, respectively. In this application, we will consider datasets for which $n_y=1$, as is standard in this area of literature. The input space is assumed to be continuous with $\mathcal{X} = \mathbb{R}^{n_x}$. 

A decision tree can be considered a specific type of graphical network. A graph $\mathcal{G}$ can be parameterised by a set of nodes $\mathcal{V}$ and edges $\mathcal{E}$, referred to as directed if the edge $(i,j)$ is an ordered pair that specifies the direction from node $\eta_i$ to node $\eta_j$. A directed graph is referred to as acyclic if there exists no sequence of edges that form a cycle (see Definition 3 in \citet{safavian1991survey}). 

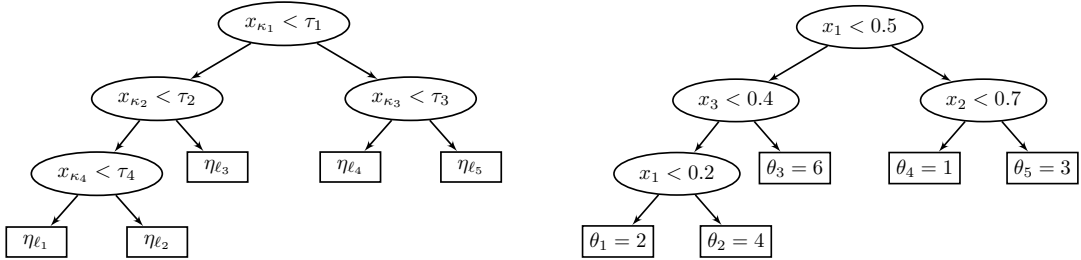
\begin{figure}[ht]
    \begin{center}
    \begin{subfigure}[t]{.49\linewidth}
        \begin{center}
            \resizebox{0.9\textwidth}{!}{
                \begin{tikzpicture}
                    \node [internal]                                        (n0)    {$x_{\kappa_1} < \tau_1$};
                    \node [internal, below=0.5cm of n0, xshift=-2.1cm]      (n1)    {$x_{\kappa_2} < \tau_2$};
                    \node [internal, below=0.5cm of n0, xshift=2.1cm]       (n2)    {$x_{\kappa_3} < \tau_3$};
                    \node [internal, below=0.5cm of n1, xshift=-1cm]        (n3)    {$x_{\kappa_4} < \tau_4$};
                    \node [leaf, below=0.5cm of n3, xshift=-1cm]            (l1)    {$\eta_{\ell_1}$};%
                    \node [leaf, below=0.5cm of n3, xshift=1cm]             (l2)    {$\eta_{\ell_2}$};
                    \node [leaf, below=0.5cm of n1, xshift=1cm]            (l3)    {$\eta_{\ell_3}$};
                    \node [leaf, below=0.5cm of n2, xshift=-1cm]             (l4)    {$\eta_{\ell_4}$};
                    \node [leaf, below=0.5cm of n2, xshift=1cm]             (l5)    {$\eta_{\ell_5}$};
                    \path [line] (n0) -- (n1);
                    \path [line] (n0) -- (n2);
                    \path [line] (n1) -- (n3);
                    \path [line] (n1) -- (l3);
                    \path [line] (n3) -- (l1);
                    \path [line] (n3) -- (l2);
                    \path [line] (n2) -- (l4);
                    \path [line] (n2) -- (l5);
                \end{tikzpicture}
            }
            \caption{Generic decision tree with defined topology $\mathcal{T}$.}
            \label{fig:ex-tree1}
        \end{center}
    \end{subfigure}
    \begin{subfigure}[t]{.49\linewidth}
        \begin{center}
            \resizebox{0.9\textwidth}{!}{
            \begin{tikzpicture}
                \node [internal]                                        (n0)    {$x_1 < 0.5$};
                \node [internal, below=0.5cm of n0, xshift=-2.1cm]      (n1)    {$x_3 < 0.4$};
                \node [internal, below=0.5cm of n0, xshift=2.1cm]       (n2)    {$x_2 < 0.7$};
                \node [internal, below=0.5cm of n1, xshift=-1cm]        (n3)    {$x_1 < 0.2$};
                \node [leaf, below=0.5cm of n3, xshift=-1cm]            (l1)    {$\theta_1 = 2$};
                \node [leaf, below=0.5cm of n3, xshift=1cm]             (l2)    {$\theta_2 = 4$};
                \node [leaf, below=0.5cm of n1, xshift=1cm]            (l3)    {$\theta_3= 6$};
                \node [leaf, below=0.5cm of n2, xshift=-1cm]             (l4)    {$\theta_4 = 1$};
                \node [leaf, below=0.5cm of n2, xshift=1cm]             (l5)    {$\theta_5 = 3$};
                \path [line] (n0) -- (n1);
                \path [line] (n0) -- (n2);
                \path [line] (n1) -- (n3);
                \path [line] (n1) -- (l3);
                \path [line] (n3) -- (l1);
                \path [line] (n3) -- (l2);
                \path [line] (n2) -- (l4);
                \path [line] (n2) -- (l5);
            \end{tikzpicture}
            }
            \caption{Decision tree with a specific parameter realisation. }
            \label{fig:ex-tree2}
        \end{center}
    \end{subfigure}
    \caption{Two decision trees with $n=4$, $n_\ell=5$, (a) generic and (b) including specific parameter values for clarity. For simplicity of exposition, the leaf nodes are parameterised by a single value.}\label{fig:ex-tree}
\end{center}
\end{figure}

A binary decision tree is a type of directed acyclic graph with the following additional properties:
\begin{itemize}
    \item There is a single node called the root node in which no edges enter. 
    \item All nodes except the root have exactly one edge that enters the node.
    \item For nodes with leaving edges, there are exactly two leaving edges.  
\end{itemize}
The following terminology is commonly used for binary decision trees: 
\begin{itemize}
    \item In each specified edge $(i,j)$, the node $\eta_i$ is referred to as the parent of the node $\eta_j$ and $\eta_j$ is referred to as the child of $\eta_i$.
    \item If a node does not have any children, it is labelled as a leaf (or terminal) node. Otherwise, it is labelled as an internal node.
    \item All nodes that have children have exactly two children, referred to as a left child $\eta_L$ and a right child $\eta_R$.
    \item The depth of a node is the number of nodes on the path from the root node to itself. 
\end{itemize}
The set of all nodes can be split into two subsets; a set of nodes that have children $\I$ and a set of terminal nodes $\Lf$, referred to as internal nodes and leaf nodes respectively. Note that with the above properties, it holds that $\I \cap \Lf = \emptyset$. The number of internal nodes is denoted by $n$ and the number of leaf nodes by $n_\ell$. Each internal node is parameterised by two variables, $\kappa$ and $\tau$, which represent the splitting index and splitting threshold respectively. The splitting index selects which data input to consider at the node, and the splitting threshold defines at what value for that input the data will divide. Each leaf node is parameterised by the variable $\theta$ required to define the assumed distribution for that leaf node. Figure \ref{fig:ex-tree1} shows an example of a generic binary decision tree with $n=4$ and $n_\ell=5$, where internal nodes are shown as ellipses and leaf nodes as rectangles. 

The parameters of a binary decision tree under the standard definition are expressed by the tuple 
\begin{equation}
    \T_\kappa = (\mathcal{T},\bm{\kappa},\bm{\tau},\bm{\theta}),
\end{equation}
where each variable denotes a specific component relating to the tree. The tree topology $\mathcal{T} = (\mathcal{V},\mathcal{E})$ holds information about the nodes $\mathcal{V}$ and edges $\mathcal{E}$ of the tree. Each tree node, $\eta$, can be further categorised either into the set of internal nodes $\I$ or the set of leaf nodes $\Lf$. As shown in Figure \ref{fig:ex-tree1}, the splitting indices are represented by $\bm{\kappa} = \left[ \kappa_1,\kappa_2,\dots,\kappa_n \right]$ and $\bm{\tau}= \left[ \tau_1,\tau_2,\dots,\tau_n \right]$ the splitting thresholds for each internal node $\eta_i \in \I$, for $i=1,\dots,n$. The parameters associated with each leaf node are $\bm{\theta} = \left[ \theta_1,\theta_2,\dots,\theta_{n_\ell} \right]$ for $\eta_{\ell_j} \in \Lf$, $j=1,\dots,n_\ell$. We define the standard decision tree with a subscript $\kappa$ here so as not to confuse with an alternative definition presented in the next section. Note that if an expression is generally applicable to either parameterisation we will drop the extra notation and simply use $\T$.

Predictions for a given input are made by traversing the tree, starting at the root node and recursively moving to one of the children nodes corresponding to the value of the relevant splitting indices for the given input, stopping once a leaf node has been reached. The output is then assumed to be associated with the distribution defined for that leaf.  Each node in the tree maintains information on the pathway taken to traverse from the root to itself. This comprises a set of ancestor nodes $\mathcal{A}(\eta)$ and a direction vector $R_\eta$. The variable $\mathcal{A}(\eta)$ denotes the set of nodes that are ancestors of node $\eta$. The vector $ R_\eta$ represents the direction of the path taken at each internal node $\eta$ while traversing the tree to the node under consideration; it takes on the value of 1 if the path goes right at node $\eta$ or 0 if it goes left. Figure \ref{fig:ex-tree2} shows a decision tree with parameters $\bm{\kappa} = [1,3,2,1]$, $\bm{\tau}=[0.5,0.4,0.7,0.2]$ and $\bm{\theta}=[2,4,6,1,3]$. Pathway information for $\eta_{\ell_2}$ is $\mathcal{A}(\eta_{\ell_2}) = [\eta_1,\eta_2,\eta_4]$ and $R_{\eta_{\ell_2}} = [ 0,0,1]$.

In Bayesian inference, probability distributions are used to quantify the uncertainty in model parameters, with the fundamental equation represented as $p(\Theta\mid \mathcal{D})$ where $\Theta$ denotes the set of model parameters and $\mathcal{D}$ refers to the dataset. For decision trees in a Bayesian context - and which is the main focus of this paper - we are interested in sampling from the distribution, 
\begin{equation}
    p(\mathbb{T}\mid \mathcal{D}).
\end{equation}
The information derived about model parameters and the uncertainty of these values can be used to define the uncertainty of the overall machine learning model predictions. This can then be used to provide confidence estimates of the output by evaluating the posterior predictive distribution. For a new datapoint $(\mathbf{x}^*,\mathbf{y}^*)$, the posterior predictive distribution can be expressed as
\begin{equation*}\label{eq:bdt-pred}
    p(\mathbf{y}^*\mid \mathbf{x}^*,\mathcal{D}) = \int p(\mathbf{y}^*\mid \mathbf{x}^*,\T)p(\T\mid \mathcal{D})d\T,
\end{equation*} 
where $p(\mathbf{y}^*\mid \mathbf{x}^*,\T)$ denotes the distribution of the decision tree output given the new input, and $p(\T\mid \mathcal{D})$ is the posterior on decision tree parameters. Monte Carlo integration techniques can be used to compute the above integral in the following manner, 
\begin{equation}\label{eq:predictive-dist-trees}
    p(\mathbf{y}^*\mid \mathbf{x}^*,\mathcal{D}) \approx \frac{1}{M}\sum_{m=1}^M p\left(\mathbf{y}^*\mid \mathbf{x}^*, \T^{m}\right), \quad  \T^{m} \sim p(\T\mid \mathcal{D})
\end{equation}
where a sample drawn from the posterior distribution on parameters is then used to compute the output of the corresponding decision tree model, which generates part of the Monte Carlo estimate.

One method to produce samples from the posterior distribution on parameters is by using a Markov chain, where each point in the chain represents a sample from the distribution \citep{chipman1998bayesian}. The Metropolis-Hastings algorithm \citep{metropolis1953equation,hastings1970monte} is commonly used to produce such a chain (see \citet{tierney1994markov} for detailed analysis of theoretical properties). Given a training dataset $\mathcal{D} = \{\X,\Y\}$ and an initial tree $\T^0$, a chain of tree samples $\T^0,\T^1,\T^2,\dots$ from the posterior distribution is generated via the Metropolis-Hastings algorithm by applying the following two steps at each iteration $i$:
\begin{enumerate}
    \item Propose a new tree $\T^*$ using transition kernel $q(\T^*\mid \T^i)$,
    \item Calculate the probability of accepting the proposed tree:
        \begin{equation}  \label{eq:accept-prob-tree}
            \alpha_{\textsc{accept}} = \min\left\{\frac{q(\T^i\mid \T^*)\ell(\mathbf{Y}\mid \mathbf{X},\T^*)p(\T^*)}{q(\T^*\mid \T^i)\ell(\mathbf{Y}\mid \mathbf{X},\T^i)p(\T^i)},1\right\}
        \end{equation}
\end{enumerate}
The proposed tree is then accepted with probability $\alpha_{\textsc{accept}}$. If accepted, the next sample is taken to be $\T^{i+1}=\T^*$, otherwise, the current tree is taken as the next sample $\T^{i+1}=\T^i$. The terms $q(\T^*\mid \T^i)$, $\ell(\mathbf{Y}\mid \mathbf{X},\T)$ and $p(\T)$ refer to the proposal distribution, likelihood distribution, and tree parameter prior respectively.

\section{RJHMC-Tree Algorithm}\label{sec:rjhmc}

As outlined in the previous section, MCMC techniques require generating new tree samples that are consistent with the data. The main difficulty with this procedure lies in defining an appropriate transition kernel $q(\T^*\mid \T^i)$ to move from the current tree sample $\T^i$ to the proposed tree $\T^*$.

There have been multiple attempts to define such kernels for decision trees that are both computationally efficient while fully exploring the posterior distribution. \citet{chipman1998bayesian} and \citet{denison1998bayesian} define their transition kernel by randomly selecting from a set of possible moves (i.e. grow/birth, prune/death, change/variable, swap), then locally changing the structure or splitting variables of selected node/s while keeping the remaining sections of the tree constant.
\citet{wu2007bayesian}, \citet{gramacy2008bayesian} and \citet{pratola2016efficient} adopt a similar approach but have included an additional possible move by which major changes are made to the tree structure and parameters (i.e. by radical restructure or rotations). \citet{taddy2011dynamic} and \citet{lakshminarayanan2013top} define their transition kernel by local proposals (i.e. grow, prune, stay) to each tree particle structure, with the selection of splitting parameters only occurring during a grow move.

A major problem with these approaches is that the proposal is often local and not informed by the likelihood, resulting in the tree infrequently moving to the newly proposed tree. The main contribution of this paper is a new transition kernel $q(\T^*\mid \T^i)$ that improves on existing methods by using information from the likelihood distribution to jointly propose new sets of parameters at each iteration. Our algorithm employs a three-part update scheme with the tree topology updated using the standard MCMC moves, but where the remaining parameters are updated using a HMC-based sampling method. The application of HMC allows the joint updating of the remaining tree parameters in a logical manner, improving on the local random-walk updates used in existing Bayesian decision tree methods. 

The other concern with MCMC-based Bayesian decision tree algorithms is the varying dimension of the parameter vector. This type of model is officially known as transdimensional, and although this may seem like an inconspicuous feature of the model at first, the implications of this require proper consideration. One such method that is designed to handle this type of model is the reversible jump HMC (RJHMC), algorithm as proposed in \citet{sen2017transdimensional}. This method considers continuous updates via HMC and explicitly accounts for changes in the dimension of the parameter vector. The underlying algorithm was first presented in \citet{al2004improving} in the context of improving the movement of reversible jump MCMC (RJMCMC) samplers. This paper suggested the inclusion of an intermediate distribution $\pi^*$, such as a tempered version of the true target distribution, to encourage movement to areas of high likelihood before the acceptance step (see \citet{neal1996sampling} for further discussion of tempering). RJHMC applies the same logic but aligns this intermediate distribution to the augmented target distribution used within the HMC algorithm. 

However, there are some aspects that are restrictive in the original RJHMC implementation. One issue, and which is true of transdimensional samplers in general, is that the current continuous variables might not make sense for the new dimension, therefore reducing the acceptance probability for that transition. This then affects the movement of the chain, where most of the time is spent rejecting samples, leading to computational inefficiencies. The acceptance probability for the RJHMC method is not immune to this pathology. In addition to this, the original RJHMC algorithm is based on an asymmetric one-sided proposal scheme, meaning that the HMC transition is only applied to the higher dimension space. However, it makes sense to also move within the lower dimension, that is, the tree with the smaller topology.

This section proposes a new algorithm, entitled RJHMC-Tree (detailed in Algorithm \ref{alg:rjhmc-tree}), which extends the method presented in \citet{sen2017transdimensional} to the decision tree model and improves the sampling routine based on the pitfalls of the current implementation. To achieve this, the asymmetric proposal scheme is extended to provide movement on both sides of a dimension jump, while also incorporating a biased acceptance probability within the burn-in of the chain to ensure quicker convergence. The augmented target distribution in HMC is used as the intermediate proposal distribution to incorporate the geometry of the likelihood to define the next proposal.

We start off this section by first presenting the two softened versions of the decision tree, after which we present the prior and likelihood definitions for both classification and regression problems. Then the overall RJHMC-Tree algorithm is presented, detailing the proposal scheme, acceptance probabilities, overall algorithm, and implemented heuristics.

\subsection{Soft decision trees}\label{sec:soft-decision-trees}
A characteristic of HMC is that it requires derivatives to define the next proposal. The standard decision tree defines a set of hierarchical splits that partition the input space into a set of disjoint regions that collectively make up the entire space. This construction requires that each datapoint $(\mathbf{x}_i, y_i )$ be associated with a single leaf node, resulting in a piecewise constant likelihood with respect to the spitting variable $\bm{\tau}$. The derivative of this likelihood is zero almost everywhere and undefined at the value of each input datapoint in a specified dimension. Consequently, HMC is incompatible under the standard decision tree construction with respect to the splitting threshold variables. 

To remove this pathology, we redefine the decision tree structure in two ways, one which involves a mix of discrete and continuous variables and the other which has a fully continuous parameterisation. First, we use `soft splits', as originally formulated in \citet{linero2018bayesian} in the context of Bayesian additive regression tree models, and extend this idea to decision trees and classification problems. This changes the decision function to consider instead the probability of each datapoint being associated with each leaf node, relaxing the hard-split constraint. Second, we define a novel `soft index' feature for decision trees by embedding the input selection at each internal node in a unit simplex. This effectively weights the different inputs at each internal node, which simplifies sampling by removing the need for discrete model parameters. We discuss both definitions now in detail, with a clear emphasis on the likelihood and prior distributions for each.

\subsubsection{Soft decision function}
The first way in which the decision tree is softened is by defining a soft approximation to the decision function; we will henceforth refer to this version as HMC soft decision function (HMC-DF). Under the soft decision function specification, the probability of a datapoint $(\mathbf{x}_i, y_i )$ being assigned to leaf node $\eta_{\ell_k} \in \Lf$ is denoted $\phi_{i,k} $. This value is calculated based on the probability of the path taken for that datapoint to reach the leaf node, defined as,
\begin{equation} \label{eq:phi}
    \phi_{i,k} (\mathbf{x}_i\mid \mathbb{T},\eta_{\ell_k}) = \prod_{\eta \in \mathcal{A}(\eta_{\ell_k})}  \psi(\mathbf{x}_i\mid \mathbb{T},\eta)^{R_{\eta}} (1-\psi(\mathbf{x}_i\mid \mathbb{T},\eta))^{1-R_{\eta}},
\end{equation}
where $\mathcal{A}(\eta)$ and $ R_\eta$  denote the set of ancestor nodes and direction vector for node $\eta$ as defined in Section~\ref{sec:model}. The probability that a datapoint goes to the left at internal node $\eta$ with splitting dimension/index $\kappa_\eta$ and threshold $\tau_\eta$ is defined as,
\begin{equation}\label{eq:psi}
    \psi(\mathbf{x}\mid \mathbb{T}_\kappa,\eta) = f\left(\frac{x_{\kappa_\eta}-\tau_\eta}{h} \right).
\end{equation}
The logistic function $f(x) = \left( 1+\exp(-x)\right) ^{-1}$ is commonly used for soft approximations of binary splits and is adopted here (but other choices can easily be accommodated). Note that in this method, the splitting index variable $\kappa$ is discrete and therefore requires a HMC-within-MH sampling approach.

\subsubsection{Soft input selection}
To further improve the utility of the HMC framework, we propose to soften the splitting input at each internal node, which removes the need for any discrete model variables. We will refer to this version as the HMC soft decision function and index (HMC-DFI). A unit simplex is used to weigh each input dimension at each internal node. That is, each internal node is now parameterised by a splitting threshold $\tau_\eta$ and splitting simplex $\Delta_\eta$ (as opposed to a splitting index $\kappa_\eta$) where,
\begin{equation}
    \Delta_\eta = \left[ w_{\eta,1}, w_{\eta,2}, \dots, w_{\eta,n_x} \right], \quad w_{\eta,i} > 0,\quad \sum_{i=1}^{n_x}w_{\eta,i} = 1.
\end{equation}
Under this new parameterisation, the decision tree is redefined as
\begin{equation}
    \T_\Delta = (\mathcal{T},\bm{\Delta},\bm{\tau},\bm{\theta}),
\end{equation}
with the subscript making the distinction clear between the two definitions. When both the standard $\mathbb{T}_\kappa$ or softened $\mathbb{T}_\Delta$ definitions are applicable based on the given context, the simplified notation $\mathbb{T}$ with the subscript dropped will be used. Equation \ref{eq:psi} is also altered to include the new soft input selection at each node,
\begin{equation}\label{eq:psi-index}
    \psi(\mathbf{x}\mid \mathbb{T}_\Delta,\eta) = f\left(\frac{\mathbf{x}\Delta_\eta-\tau_\eta}{h} \right).
\end{equation}
Note the clear distinction between when the full input vector for each datapoint is being used (i.e. via $\mathbf{x}$), and when only a single input is selected (i.e. $x_{\kappa_\eta}$).

\subsection{Likelihood Definitions}\label{sec:likelihoods}

Both likelihood definitions are directly related to Equation \ref{eq:phi}, which is generally applicable, where the distinction between the two methods is made through the definition of $\psi$. Under the following assumptions and definitions, the derivatives of the likelihood with respect to the continuous tree parameters have been computed, with the analytical expressions presented in Appendix \ref{sec:app-llh}.

\subsubsection{Classification}

The likelihood for soft classification trees  is taken to be the Dirichlet-Multinomial joint compound as defined in Equation \ref{eq:llh-class},
\begin{equation}\label{eq:llh-class}
    \ell(\mathbf{Y}\mid \mathbf{X},\T) = \prod_{k=1}^{n_\ell} \left[ \frac{\Gamma(A)}{\Gamma(\Phi_k+A)} \prod_{m=1}^M\frac{\Gamma(\bm{\phi}_{m,k} +\alpha_m)}{\Gamma(\alpha_m)} \right]
\end{equation}
where $M$ refers to the number of output classes, $n_\ell$ is the number of leaf nodes in the tree and $\Phi_k = \sum_m \bm{\phi}_{m,k}$ for each leaf $\eta_{\ell_k}$. The variable $\bm{\phi}_{m,k} $ represents the probability of each datapoint $(\mathbf{x}_i,y_i)$, for which the output class is $y_i = m$, being assigned to leaf node $\eta_{\ell_k}$ in tree $\T$ and is expressed as
\begin{equation}\label{eq:temp}
    \bm{\phi}_{m,k}  = \sum_{i=1}^N\phi_{i,k}(\mathbf{x}_i\mid \T,\eta_{\ell_k})\mathbb{I}(y_i = m)
\end{equation}
where $\phi_{i,k}$ is as previously defined in Equation~\ref{eq:phi}. Note that other tree parameters (i.e. $\tau_\eta$, $\kappa_\eta$/$\Delta_\eta$) are included in the above expression through calculation of $\phi_{i,k}$  which depends on the definition of $\psi$, but have been removed for notational simplicity. Note that the likelihood for soft classification trees differs from the standard hard-split likelihood used in literature through the terms $\Phi_k$ and $\bm{\phi}_{m,k}$, which denote probabilities for each datapoint (thus with a value in [0,1]) as opposed to an integer denoting a discrete number of datapoints.

\subsubsection{Regression}

 Following \citet{linero2018bayesian}, the likelihood for regression trees in the soft setting is defined to be, 
\begin{equation}\label{eq:llh-reg}
    \ell\left(\mathbf{Y} \mid \mathbf{X},\T\right) = \prod_{i=1}^N\left(2\pi\sigma^2\right)^{-\frac{1}{2}}\exp\left[ -\frac{1}{2\sigma^2} \left(  \sum_{k=1}^{n_\ell} \phi_{i,k}(\mathbf{x}_i)\cdot(\mu_k - y_i) \right) ^2 \right],
\end{equation}
where $\phi_{i,k}$ is as defined in Equation~\ref{eq:phi}, $\mu_k$ is the mean value of the assumed normal distribution within each leaf node $\eta_{\ell_k}$ and $\sigma$ is the assumed constant variance across all leaf nodes. Note that again, tree parameters such as $\tau_\eta$, $\kappa_\eta$, or $\Delta_\eta$ are implicitly included through the calculation of $\phi_{i,k}$. The difference to the standard hard-split likelihood used in Bayesian regression trees again comes through the term $\phi_{i,k}$ which represents the probability of each datapoint in the leaf node, whereas in the hard-split case, this variable exists instead as an indicator function for each datapoint in each leaf node.

\subsection{Prior Specification}\label{sec:prior}

The prior definition can be split in a hierarchical manner whereby first the tree topology is drawn from a topology prior, then the remaining parameters are drawn independently from their respective priors with the dimension dependent on the number of internal and leaf nodes. The prior definitions for the soft decision function and soft index are shown in Equations \ref{eq:tree-prior-soft-1} and \ref{eq:tree-prior-soft-2} respectively,
\begin{subequations}
    \begin{equation} \label{eq:tree-prior-soft-1}
        p(\T_\kappa) = p_\tau(\bm{\tau}\mid \mathcal{T})  p_\kappa(\bm{\kappa}\mid \mathcal{T}) p_\theta(\bm{\theta}\mid \mathcal{T}) p(\mathcal{T}),
    \end{equation}
    \begin{equation} \label{eq:tree-prior-soft-2}
        p(\T_\Delta) = p_\tau(\bm{\tau}\mid \mathcal{T})  p_\Delta(\bm{\Delta}\mid \mathcal{T}) p_\theta(\bm{\theta}\mid \mathcal{T}) p(\mathcal{T}).
    \end{equation}
\end{subequations}
We assume the following priors on parameters by default. Note that these are defined to ensure the flexibility of the model, as opposed to efficiency through conjugacy as implemented by previous methods.
\begin{align*}
    \kappa_i &\sim \mathbb{C}(\bm{p}), \quad \Delta_i \sim \text{Dir}(\bm{\alpha}), \quad \tau_i \sim \mathcal{B}(1,1), && \eta_i \in \I, \quad i=1,\dots,n \\
    \mu_j &\sim \mathcal{N}(\alpha_\mu,\beta_\mu), \quad \sigma  \sim \Gamma^{-1}(\alpha_\sigma,\beta_\sigma), && \eta_{\ell_j} \in \Lf,\quad j=1,\dots,n_\ell
\end{align*}
where $\mathbb{C}$ denotes the categorical distribution, Dir the Dirichlet distribution, $\mathcal{B}$ the beta distribution, $\mathcal{N}$ the normal distribution and $\Gamma^{-1}$ the inverse-gamma distribution. For the categorical distribution, each probability hyperparameter is set to $p_i=\frac{1}{n_x}$, the Dirichlet concentration parameters are set to $\alpha_i=1$, the default hyperparameters on the normal distribution are set to be $\alpha_\mu=0$ and $\beta_\mu=1$, and a constant variance model is assumed for regression trees with $\alpha_\sigma = \beta_\sigma = \frac{3}{2}$. Note that inputs are transformed into the unit interval such that the prior on the splitting threshold is uniform over the range of each dimension.

The prior on the decision tree topology is taken as the standard used in literature, as originally defined in \citet{chipman1998bayesian}, with the definition provided in Equation \ref{eq:cgm-prior},
\begin{equation}\label{eq:cgm-prior} 
    p(\mathcal{T}) \propto \prod_{\eta_i \in \I} p_{\textsc{split}}(\eta_i) \times \prod_{\eta_{\ell_j} \in \Lf} (1-p_{\textsc{split}}(\eta_{\ell_j})),
\end{equation}
where $p_{\textsc{split}}(\eta)$ denotes the probability that a node $\eta$ will split. This was originally defined as per Equation \ref{eq:cgm-prior-prob},
\begin{equation}\label{eq:cgm-prior-prob}
    p_{\textsc{split}} = \alpha(1+d_\eta)^{-\beta},
\end{equation}
where $\alpha \in (0,1)$ and $\beta \geq 0$ are hyperparameters and where $d_\eta$ represents the depth of the node in the tree. 

\subsection{RJHMC-Tree}

This section will discuss the proposal scheme and acceptance probabilities of the RJHMC-Tree algorithm in detail. The overall algorithm is then presented, along with additional practical heuristics that have been found to improve performance.

\subsubsection{Proposal Scheme}

The RJHMC-Tree algorithm proposed in this section (see Algorithm \ref{alg:rjhmc-tree}) employs a three-part update scheme that separates the problem into a possible change in the tree topology that occurs between updates to the remaining tree parameters. The transition kernel can be broken down as follows,
\begin{multline}
    q\left(\T^* \mid \T^i\right) 
    = \underbrace{q_\textsc{hmc}\left(\mathcal{T}^*,\bm{\kappa}^*,\bm{\tau}^*,\bm{\theta}^*\mid \mathcal{T}^*,\bm{\kappa}'',\bm{\tau}'',\bm{\theta}''\right)}_{\mathclap{\text{ HMC-based transition}}} \times \\
    \underbrace{q_{mn}\left(\mathcal{T}^*,\bm{\kappa}'',\bm{\tau}'',\bm{\theta}'' \mid \mathcal{T}^i,\bm{\kappa}',\bm{\tau}',\bm{\theta}' \right)}_{\mathclap{\text{transdimensional transition}}} \times \\
    \underbrace{q_\textsc{hmc}\left(\mathcal{T}^i,\bm{\kappa}',\bm{\tau}',\bm{\theta}' \mid \mathcal{T}^i,\bm{\kappa}^i,\bm{\tau}^i,\bm{\theta}^i\right)}_{\mathclap{\text{HMC-based transition}}}
\end{multline}
where $q_\textsc{hmc}$ denotes a HMC-based transition and $q_{mn}$ a transition between dimensions. The proposal scheme first updates the tree parameters in the current dimension using a HMC-based transition, then updates the topology $\mathcal{T}$ of the decision tree using one of the standard moves (i.e. grow, prune, stay) before again updating the remaining parameters. The model parameters are jointly updated in either a HMC-within-MH or purely HMC manner via the use of soft approximations to the standard decision tree as defined in Section \ref{sec:soft-decision-trees}.

The transdimensional step defines the transition between different tree topologies. The assumed independence between tree topology and parameters implies that this transition kernel can be simplified as follows:
\begin{equation}
    \mathcal{T}^* \sim q_{mn}\left(\mathcal{T}^*,\bm{\kappa}'',\bm{\tau}'',\bm{\theta}'' \mid \mathcal{T}^i,\bm{\kappa}',\bm{\tau}',\bm{\theta}' \right) = q_{mn}\left(\mathcal{T}^* \mid \mathcal{T}^i\right)
\end{equation}
with the topological change defined as either
\begin{itemize}
    \item $q_\textsc{grow}$: one leaf node $\eta_{\ell_k} \in \Lf$ is selected at random to be grown (i.e. converted to an internal node by adding two children nodes),
    \item $q_\textsc{prune}$: one node in which both children nodes are leaves (or terminal) is selected at random to be pruned (i.e. removed from the tree),
    \item $q_\textsc{stay}$: the topology does not change.
\end{itemize}

The probability of each topology move is a hyperparameter defined by the user and can be changed depending on whether the method is in the warm-up phase or sampling phase.

Given a specific topology, the remaining tree parameters are updated using either HMC-within-MH or pure HMC, relating to the HMC-DF and HMC-DFI methods respectively (see Section \ref{sec:soft-decision-trees}). This HMC-based update occurs at both the start and end of the overall transition and is defined as either
\begin{itemize}
    \item $q_\textsc{hmc-df}$: a new set of splitting indices $\bm{\kappa}$ are randomly proposed via MH, and given this, the continuous parameters $\bm{\tau}$ and $\bm{\theta}$ are jointly updated via HMC,
    \item $q_\textsc{hmc-dfi}$: all parameters $\bm{\Delta}$, $\bm{\tau}$ and $\bm{\theta}$ are jointly updated via HMC.
\end{itemize}

The RJHMC-Tree algorithm is based on the assumption that the dimension of the parameter vector varies throughout the sampling routine. As such, a main feature of the transition kernel is that it acts on a parameter space that changes size with a change in the tree dimension. This requires that the second HMC-based transition be commenced at the previously accepted parameter values. That is, after the topology transition has been proposed (i.e. either grow, prune or stay), the parameter values are started at the values of the previous tree, or, in the case that the node did not exist in the previous tree, the parameters are drawn from their respective prior distributions. See Appendix \ref{sec:rjhmc-tree-move} for further details.

\subsubsection{Acceptance Probabilities}\label{sec:accept-probs}

Exploration of the Bayesian decision tree posterior via RJHMC-Tree can be thought of as generating samples from a transdimensional model, where one parameter dictates the submodel size (via the topology variable) and subsequently the dimensions of the other parameters to be sampled. That is, the parameter vector will necessarily vary across the sequence of samples, resulting in a parameter space that changes dimension.

A major issue integral to all transdimensional samplers is that the acceptance probability corresponding to the change in the discrete variable can be very low. This causes issues with convergence when either the input dimension or the underlying tree is sufficiently large. It would therefore make sense to assist the initial movement of the chain in order to converge quicker to an area of high likelihood before generating samples from the posterior distribution. As a result, two acceptance probabilities are employed within the RJHMC-Tree algorithm. 

The main acceptance probability is an extension of that presented in \citet{sen2017transdimensional} and is applied during the second part of the sampling routine. Equation \ref{eq:rjhmc-tree-accept-prob} shows this expression, with the derivation provided in Appendix \ref{sec:rjhmc-tree-move},
\begin{equation} \label{eq:rjhmc-tree-accept-prob}
    \alpha_{\textsc{accept,rj}} = \min\left\{\frac{\pi_n(\mathbf{x}^*)\pi_n^*(\mathbf{x}'')\pi_m^*(\mathbf{x})q_{nm}(\mathbf{x}'|\mathbf{x}'')}{\pi_m(\mathbf{x})\pi_n^*(\mathbf{x}^*)\pi_m^*(\mathbf{x}')q_{mn}(\mathbf{x}''|\mathbf{x}')},1\right\}.
\end{equation}
The other is a biased version which is used during the burn-in phase, allowing the chain to converge rapidly to areas of high probability while targeting areas similar to the true distribution. This simplifies the acceptance probability to 
\begin{equation}\label{eq:rjhmc-burn-in-accept}
    \alpha_{\textsc{burn-in}} = \min\left\{\frac{\pi_n(\mathbf{x}^*)q_{nm}(\mathbf{x}'|\mathbf{x}'')}{\pi_m(\mathbf{x})q_{mn}(\mathbf{x}''|\mathbf{x}')},1\right\},
\end{equation}
where the ratios corresponding to the intermediate distributions are removed. Note that these ratios are the main parts of the acceptance probability defined in Equation \ref{eq:rjhmc-tree-accept-prob} that cause issues with movement. The biased acceptance probability therefore contains only the ratio of the initial and final likelihoods and the proposal distributions. 

\subsubsection{Overall Algorithm}

Algorithm \ref{alg:rjhmc-tree} details the overall RJHMC-Tree algorithm for both HMC-DF and HMC-DFI. The first set of intermediate tree parameters is determined via either HMC-within-MH or HMC in steps 1a or 1b respectively. A new tree topology is proposed in steps 2-3 with any additional parameters required for the reversible jump drawn from the respective priors in step 4 (see Appendix \ref{sec:rjhmc-tree-move} for further details). Next, the tree parameters corresponding to the new tree topology are determined, again via either HMC-within-MH or HMC in steps 5a or 5b respectively. The proposed tree is accepted in step 6 based on the current iteration, with acceptance probabilities as defined in Section \ref{sec:accept-probs}.
This sampling method is applied during the entire simulation of the chain, with a change in acceptance probability after the burn-in phase, to produce a final sequence of tree samples $\{\T^i\}_{i=0}^N$.

\begin{algorithm}[hbtp]
    \DontPrintSemicolon
    \SetAlgoLined
    \SetKwComment{tcb}{(}{)}
    \KwIn{Number of sampling iterations $N$, initial tree $\T^0$, number of HMC samples to take through intermediate distributions $k$, number of burn-in iterations $N_\textsc{burn-in}$}
    \For{$i = 0 $ \KwTo $N$}{
    1a. Run $k$ HMC-within-MH steps for the current decision tree $\T_\kappa^i$ via kernel $q_{\textsc{hmc-df}}$ to arrive at $\left(\bm{\kappa}',\bm{\tau}',\bm{\theta}'\right)$ \;
    1b. Run $k$ HMC steps for the current decision tree $\T_\Delta^i$ via kernel $q_{\textsc{hmc-dfi}}$ to arrive at $\left(\bm{\Delta}',\bm{\tau}',\bm{\theta}'\right)$ \;
    2. Draw $C \sim$ Categorical$(\bm{p_m})$ where $\bm{p_m}=\left[ p_{\textsc{grow}},p_{\textsc{prune}},p_{\textsc{stay}} \right]$ \;
    \uIf{$C=\textsc{grow}$}{
        $q_{mn} = q_\textsc{grow}$\;} 
    \uElseIf{$C=\textsc{prune}$}{
        $q_{mn} = q_\textsc{prune}$\;} ;
    \Else(\tcb*[h]{$C=\textsc{stay}$}){
        $q_{mn} = q_\textsc{stay}$\; }
    3. Draw a new tree topology from the corresponding transition kernel $ \mathcal{T}^* \sim q_{mn}\left(\mathcal{T}^* \mid \mathcal{T}^i\right)$. \;
    4. Draw new parameters as required for the reversible jump step. \;;
    5a. Run $k$ HMC-within-MH steps for the proposed decision tree $\T_\kappa^*$ via kernel $q_{\textsc{hmc-df}}$ to arrive at $\left(\bm{\kappa}^*,\bm{\tau}^*,\bm{\theta}^*\right)$ \;
    5b. Run $k$ HMC steps for the proposed decision tree $\T_{\Delta}^*$ via kernel $q_{\textsc{hmc-dfi}}$ to arrive at $\left(\bm{\Delta}^*,\bm{\tau}^*,\bm{\theta}^*\right)$ \;
    6. Accept the final proposal with probability $\alpha$ where,\;
    \uIf{$ i < N_\textsc{burn-in}$}{
        $\alpha = \alpha_{\textsc{burn-in}}$ (see Equation \ref{eq:rjhmc-burn-in-accept}). \;
    } 
    \Else{
        $\alpha = \alpha_{\textsc{accept,rj}}$ (see Equation \ref{eq:rjhmc-tree-accept-prob}). \;}
    }
    \KwOut{Sequence of decision trees $\{\T^i\}_{i=0}^N$.}
    \caption{RJHMC-Tree Algorithm}
    \label{alg:rjhmc-tree}
\end{algorithm}

\subsubsection{Practical Implementation}
Multiple heuristics have been implemented in the sampling routine to help improve performance. This section describes the additions that have been found to improve the mixing of chains in practice. 

\subsubsection*{Intermediate Burn-in}\label{sec:rjhmc-nuts}

When exploring the posterior of the decision tree model there is a possible change in dimension at every sampling iteration, i.e. by selecting a new tree structure. Not only may this change the optimal value of the parameters, but in the case of HMC, it could also change the optimal value of the sampling method hyperparameters. The HMC algorithm has multiple tuning hyperparameters that assist in the sampling method, with the main two being the step size and mass matrix. The specific HMC implementation used is the No-U-Turn (NUTS) sampler (see \citet{hoffman2014no} for details) which adapts the hyperparameters of the method while moving through the intermediate distributions to account for the possible change in model dimension. 

For the HMC-DF method, the MH step updates each splitting index by first randomly permuting the order in which the values are updated, then by selecting from one of the dimensions at random, choosing to accept or reject the individual update at each stage using the standard acceptance probability \citep{metropolis1953equation,hastings1970monte}. The continuous parameters are then updated using NUTS, whereby the proposed sample is arrived at by the simulation of HMC trajectories. This updating method has been implemented using \textsc{NumPyro} \cite{phan2019composable} which implements the algorithm described in \citet{liu1996peskun}. For the HMC-DFI method, all parameters are continuous and therefore are updated via NUTS.

\subsubsection*{Adapting softness of splits}
We have found that varying the split sharpness parameter $h$ (see Equation \ref{eq:psi} and Equation \ref{eq:psi-index}) during the NUTS adaption phase allows the chain to converge to areas of high likelihood while still maintaining the efficiency of the remaining NUTS adaptation parameters. In practice, we make use of the adaptation windowing schedule used in both \textsc{NumPyro} \citep{phan2019composable} and \textsc{Stan} \citep{stan2019} and adapt the split sharpness parameter linearly with respect to the warm-up iteration $j$ and current window $w$ as follows:
\begin{equation}
    h=
    \begin{cases}
        h_\textsc{init} & \text{for}\quad j < w_1  \\
        h_\textsc{init} + j\times\frac{h_\textsc{final}-h_\textsc{init}}{\ell_w} & \text{for} \quad w_1 \leq j \leq w_{-2} - 100 \\
        h_\textsc{final} & \text{for}\quad j > w_{-2} - 100
    \end{cases}
\end{equation}
with hyperparameters $h_\textsc{init}$ and $h_\textsc{final}$ to be specified and where $\ell_w = w_{-2} - w_{1} + 100$. Note that as $h\to0$ we recover the original decision tree definition. 
Refer to Section~\ref{sec:discussion} for further discussion of these hyperparameters.

\section{Experiments}\label{sec:results}
We tested our method on both synthetic and real-world datasets, and compared to the MCMC-based methods presented in \citet{chipman1998bayesian} and \citet{wu2007bayesian}, and the SMC method in \citet{lakshminarayanan2013top}. Each method was tested by running the algorithm 10 times, restarting each time under a different seed value. See Table \ref{tab:grid-search} and Table \ref{tab:hyperparams} in Appendix~\ref{sec:app-hyp} for information about the grid search and final hyperparameters used for each method, respectively.

The primary metric used for comparison between methods was pointwise predictive accuracy as defined by either misclassification or mean squared error (MSE) on a testing dataset for classification or regression problems respectively \citep{gelman1995bayesian}. We were unable to compare via log predictive density due to the changing model definitions (and therefore likelihood) between methods. Furthermore, the effective sample size metric is inapplicable due to the transdimensional nature of this problem. Additional comparisons are made with respect to tree complexity, as measured by the number of leaf nodes in the tree, and movement around the parameter space, as characterised by the number of move acceptances. 

Metrics for the MCMC-based methods were calculated using samples after the warm-up phase and averaged across the 10 restarts, except the acceptance rate which was calculated over the whole chain. For SMC, training and testing metrics corresponding to the final iteration (using all particles, weighted appropriately) were used and again averaged across each of the 10 restarts. The average number of leaves was calculated as the average of all particle trees at each iteration. Certain metrics may not be applicable to a particular type of sampling method, i.e., the number of move acceptances for SMC methods. Similarly, some methods are not applicable due to restrictions in their implementation (i.e. not applicable to regression datasets or more than two classification classes). In each case, this has been either noted or the method not included in the comparison tables below.

In the following sections, we will denote the method presented in \citet{chipman1998bayesian} by CGM, the method in \citet{lakshminarayanan2013top} by SMC and that in \citet{wu2007bayesian} by WU. Our methods will be denoted HMC-DF and HMC-DFI as discussed in Section \ref{sec:rjhmc}. We note that a single iteration in the WU method corresponds to 25 individual change, grow/prune and swap propose-accept steps, one restructure propose-accept, then another 25 individual change, grow/prune and swap propose-accept steps (i.e. includes 50 standard CGM iterations in one). For this reason, we present both the per proposal and per iteration (shown in brackets) acceptance rates in the comparison tables below. 

\subsection{Simulated data}
\begin{figure}[ht]
    \begin{center}
        \resizebox{0.4\textwidth}{!}{
        \begin{tikzpicture}
            \node [internal]                                        (n0)    {$x_1 < 4$};
            \node [internal, below=1cm of n0, xshift=-2cm]      (n1)    {$x_0 < 3$};
            \node [internal, below=1cm of n0, xshift=2cm]       (n2)    {$x_0 < 5$};
            \node [leaf, below=1cm of n1, xshift=-1cm]        (n3)    {$\mu=1$};
            \node [internal, below=1cm of n1, xshift=1cm]            (l3)    {$x_0 < 7$};
            \node [leaf, below=1cm of n2, xshift=-1cm]             (l4)    {$\mu=8$};
            \node [leaf, below=1cm of n2, xshift=1cm]             (l5)    {$\mu=2$};
            \node [leaf, below=1cm of l3, xshift=-1cm]             (l6)    {$\mu=5$};
            \node [leaf, below=1cm of l3, xshift=1cm]             (l7)    {$\mu=8$};
            \path [line] (n0) -- (n1);
            \path [line] (n0) -- (n2);
            \path [line] (n1) -- (n3);
            \path [line] (n1) -- (l3);
            \path [line] (n2) -- (l4);
            \path [line] (n2) -- (l5);
            \path [line] (l3) -- (l6);
            \path [line] (l3) -- (l7);
        \end{tikzpicture}
        }
        \caption{True tree definition used to generate the synthetic dataset adapted from \citet{chipman1998bayesian}. }
        \label{fig:cgm-tree}
    \end{center}
\end{figure}
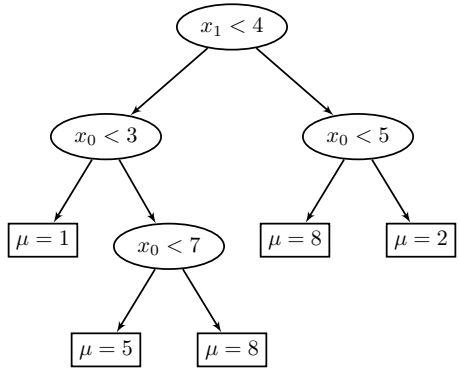

The synthetic dataset, originally presented in \citet{chipman1998bayesian} and denoted here by CGM, was used to confirm the validity of our method and compare against other Bayesian decision tree methods. The dataset was adapted such that all inputs are numerical with $n_x = 2$ and simulated via the tree structure shown in Figure \ref{fig:cgm-tree} to give $N_\textsc{train} = 800$ training and $N_\textsc{test}=800$ testing datapoints. The variance is assumed constant across nodes with value $\sigma^2 = 0.2^2$. Note that this is marginally different from the original version, where here the first split is defined by a continuous input, as opposed to categorical as in the original definition. HMC-based and WU methods were run for 1000 iterations, assuming a warm-up phase of 500 iterations. The CGM method was run for 5000 iterations to ensure sufficient burn-in, taking the last 500 iterations to calculate comparison metrics. 

\begin{figure}[ht]
  \centering
   \includegraphics[width=0.9\textwidth]{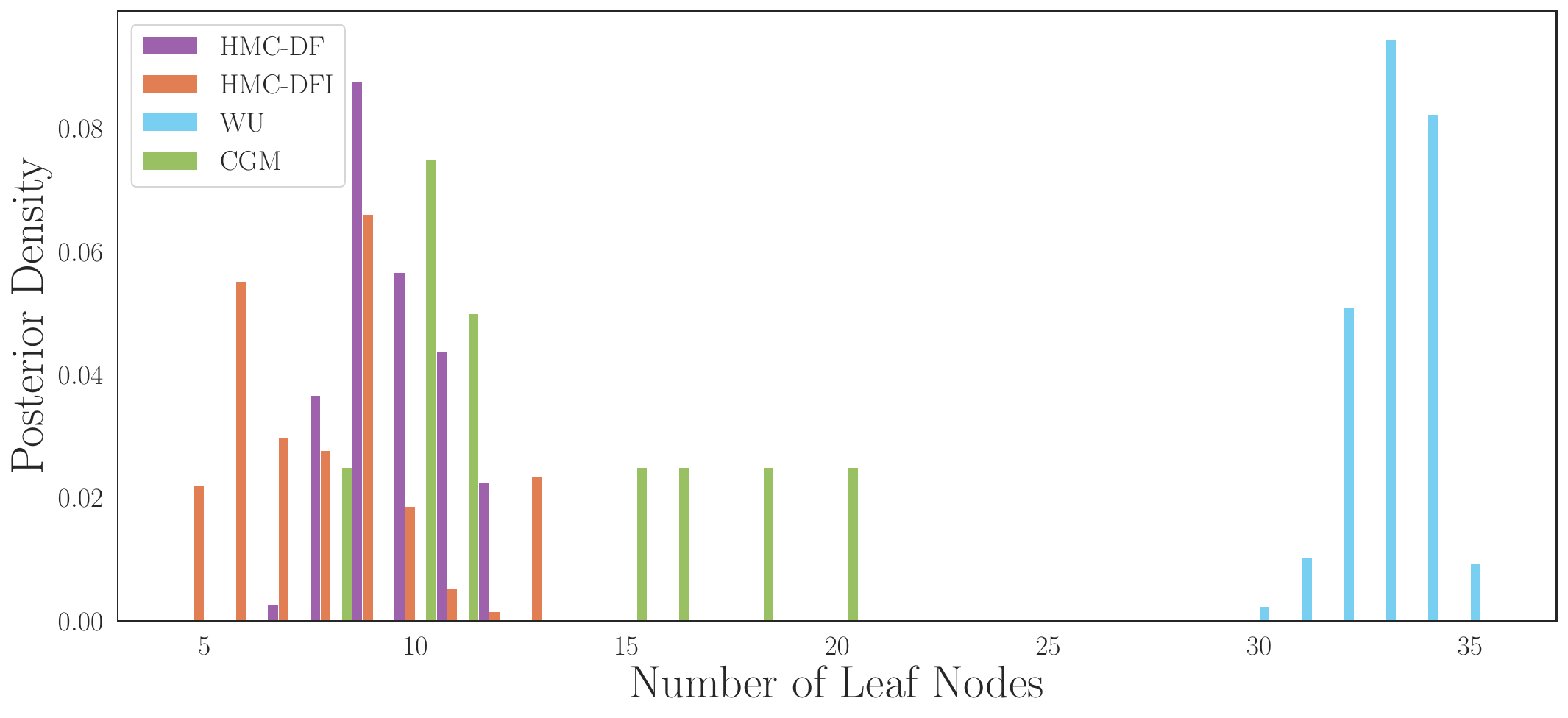}
  \caption{Comparison of the number of leaf nodes between methods on the synthetic dataset adapted from  \citet{chipman1998bayesian}. Note that the true number of leaves is $n_\ell =  5$.}
  \label{fig:cgm-nl}
\end{figure}

One of the difficulties of applying Bayesian inference to decision trees is determining the correct model dimension. Figure \ref{fig:cgm-nl} shows the distribution of the number of leaf nodes for each method after the burn-in phase for the CGM dataset.   
We can see that the highest posterior density for both HMC-based methods is skewed closer towards the correct number of leaf nodes than that of other methods. On average, the CGM method slightly overestimates the number of leaf nodes (this is consistent with the results of the original paper). The WU method significantly overestimates the number of leaf nodes, with the majority of the posterior density sitting on values of $n_\ell=30$ or greater.

\begin{figure}[ht]
\begin{center}
\includegraphics[width=0.9\textwidth]{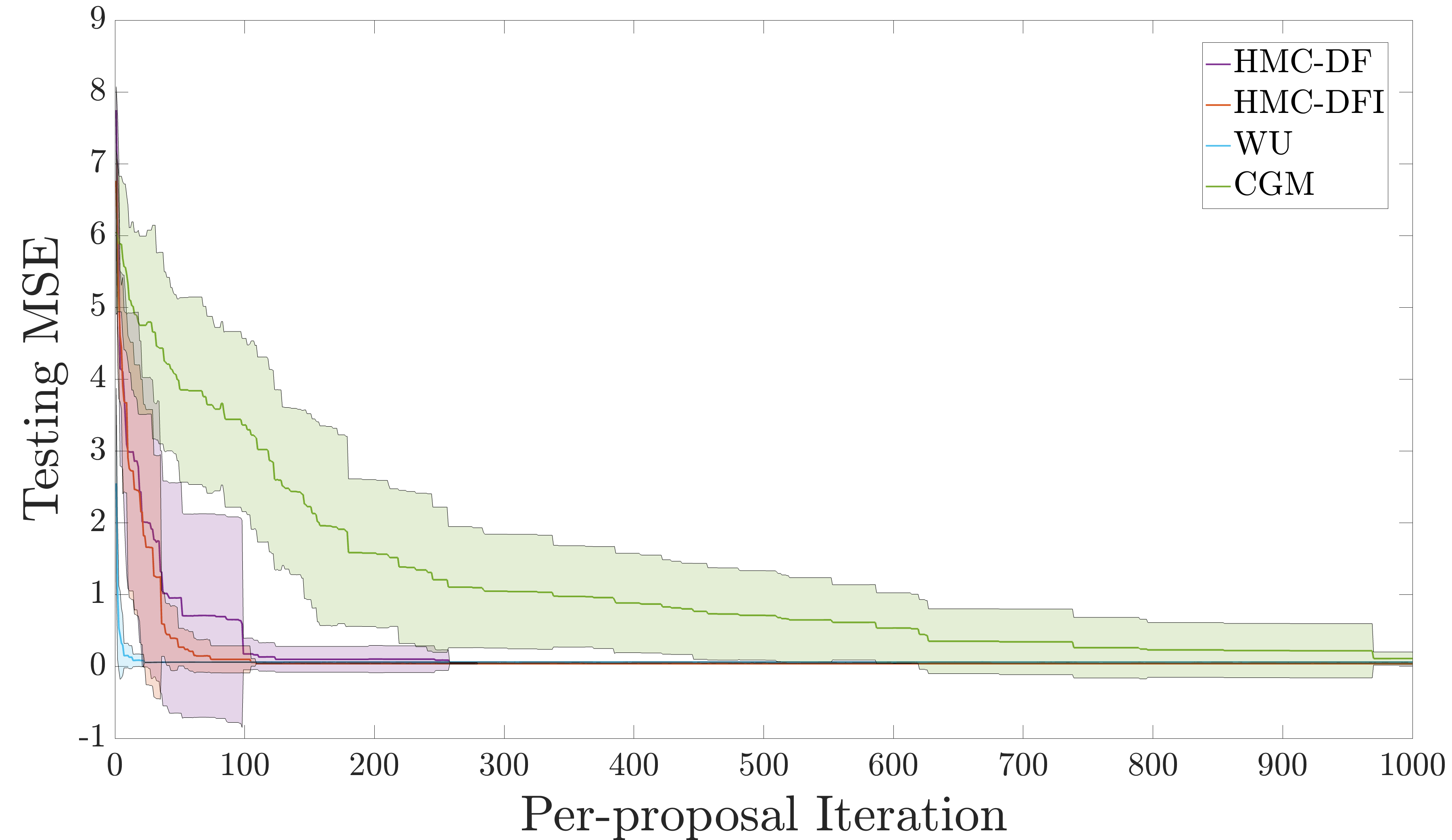}
  \caption{Comparison of the predictive error on a testing dataset between methods on the synthetic dataset adapted from \citet{chipman1998bayesian}. The x-axis represents the per-proposal iteration. The solid line denotes the mean value across the 10 chains, and the shaded area is one standard deviation.}
  \label{fig:cgm-metrics}
  \end{center}
\end{figure}

The benefit of using likelihood-based proposals can also be observed in Figure \ref{fig:cgm-metrics}. Both HMC-based methods converge in significantly fewer proposals than the standard random-walk MCMC method developed in \citet{chipman1998bayesian}. Note that the HMC-based methods have multiple HMC iterations within a single proposal iteration so will be more complex to implement than \citet{chipman1998bayesian} which may explain the faster per-proposal iteration convergence. The WU method converges in fewer proposal iterations again, however, one iteration includes 50 standard CGM propose-accept iterations as noted previously.

Although the WU method converges in fewer iterations, further comparison of metrics shows that it is overfit to the training dataset - as it gives the best training accuracy but comparatively worse testing error. Table \ref{tab:synth-metrics} presents the training and testing accuracy, average number of leaf nodes, and acceptance rate for the different methods applied to the CGM dataset. The standard deviation of the performance metrics is included in parentheses and the per iteration acceptance rate for the WU method is shown in brackets. Note that the SMC method has not been included here as its implementation is not applicable to regression datasets. The best-performing method for each metric is noted in bold.

\begin{table}[H]
    \caption{Comparison of various methods for the CGM dataset. }
    \label{tab:synth-metrics}
    \centering
    \begin{scshape}
    \begin{small}
        \resizebox{0.9\textwidth}{!}{
            \begin{tabular}{lrcccccc}
            \toprule
             & & CGM  & WU & HMC-DF & HMC-DFI \\
            \midrule
             \multirow{4}{*}{CGM}  &  Train MSE & 0.043(1.9e-4)  &  \textbf{0.042(5.8e-5)}  &  0.043(2.0e-4)  & 0.043(1.9e-4) \\
             &  Test MSE  & 0.064(0.014)  &   0.062(0.001) & 0.041(4.6e-4) & \textbf{0.041(4.5e-4)} \\
             &  Ave. Leaves  &  12.90  & 33.1 &  9.68  & \textbf{8.15} \\
             &  Acpt. Rate  & 0.02 & 2.17[\textbf{95.9}] & 36.34  & \textbf{36.45} \\
            \bottomrule
            \end{tabular}
        }
    \end{small}
\end{scshape}
\end{table} 

A significant improvement in the per-proposal acceptance rate is shown when using HMC-based methods. Both HMC-DF and HMC-DFI have the best recorded predictive accuracy on the testing data across the 10 chains (with HMC-DFI only just better) while also having the lowest tree complexity as measured by the number of leaf nodes.

\subsection{Real-world datasets}

Our method was tested and compared against multiple real-world datasets: Iris,  Breast Cancer Wisconsin (Original), Wine, and Raisin Datasets, all of which are available from the UCI Machine Learning Repository \citep{Dua:2019}. 

Table \ref{tab:real-metrics} presents the results for each method on the real-world datasets. The predictive accuracy on both the training and testing data is reported as the average value after burn-in across the 10 chains with the standard deviation shown in parentheses. The percentage acceptance rate for the method across the entire chain and the tree complexity as measured by the number of leaf nodes are also presented.  Methods whose implementation is not compatible with the dataset have been noted as such. The best-performing method for each metric is noted in bold.

\begin{table}[ht]
    \caption{Comparison of metrics for various methods for different real-world datasets.}
    \label{tab:real-metrics}
    \centering
    \begin{scshape}
    \begin{small}
        \resizebox{\textwidth}{!}{
    \begin{threeparttable}
        \begin{tabular}{lrccccccc}
            \toprule
             & & CGM  & SMC & WU & HMC-DF & HMC-DFI \\
            \midrule
             \multirow{4}{*}{BCW}  &  Train Acc. & 0.983(0.004)  & \textbf{0.987(0.004)} & 0.978(0.005) & 0.973(0.005) & 0.981(0.003)  \\
             &  Test Acc.  &  0.939(0.014)  &  0.924(0.010) & 0.922(0.017)  & 0.940(0.010) & \textbf{0.952(0.007)} \\
             &  Ave. Leaves  &   16.55  & 18.55 & 5.09 & 5.05 & \textbf{4.17} \\
             &  Acpt. Rate  & 36.47 & N/A & 6.58[\textbf{100}] & 54.09 & \textbf{58.64} \\
            \midrule
            \multirow{4}{*}{Iris}    & Train Acc. &  \textbf{0.985(0.007)}  & 0.981(0.004)  & \multirow{4}{*}{N/A\tnote{1}}  & 0.977(0.010) & 0.975(0.009) \\
                &  Test Acc.  & 0.908(0.022)  &  0.909(0.022) &   &  0.906(0.026) & \textbf{0.917(0.023)} \\
               &  Ave. Leaves  & 4.33  & 4.55  &  & \textbf{3.38} & 3.45 \\
               & Acpt. Rate   & 8.24 & N/A &  & 56.92 &  \textbf{59.01} \\
            \midrule
            \multirow{4}{*}{Wine}    & Train Acc. & 0.957(0.016) & \textbf{0.985(0.011)}  &  \multirow{4}{*}{N/A\tnote{1}} & 0.949(0.021) & 0.952(0.016) \\
                &  Test Acc.  & 0.916(0.046)  &  \textbf{0.978(0.022)} &  & 0.950(0.039) & 0.948(0.022)\\
               &  Ave. Leaves  & 8.92  &  9.20 &  & \textbf{5.92} & 6.12 \\
               & Acpt. Rate   & 19.64  & N/A &  & 53.04 & \textbf{53.11} \\
            \midrule   
            \multirow{4}{*}{Raisin}    & Train Acc. & 0.864(0.007)  &  0.863(0.004) & 0.862(0.007)  & \textbf{0.866(0.003)} & 0.864(0.005) \\
                &  Test Acc.  & 0.843(0.010)   & 0.842(0.010)  & 0.843(0.012)  & \textbf{0.847(0.004)} &  0.838(0.007) \\
               &  Ave. Leaves  & 5.00   & 4.58  & 3.59  & \textbf{3.25} & 3.50 \\
               & Acpt. Rate   & 10.16 & N/A &  6.70[\textbf{99.84}]  & \textbf{55.53} & 53.62 \\
            \bottomrule
            \end{tabular}
    \begin{tablenotes}\footnotesize
    \item[1] Uses Binomial likelihood (only two output classes allowed)
    \end{tablenotes}
    \end{threeparttable}
        }
\end{small}
\end{scshape}
\end{table}  

\begin{figure}[H]
    \begin{subfigure}[t]{.49\textwidth}
      \centering
      \includegraphics[width=\linewidth]{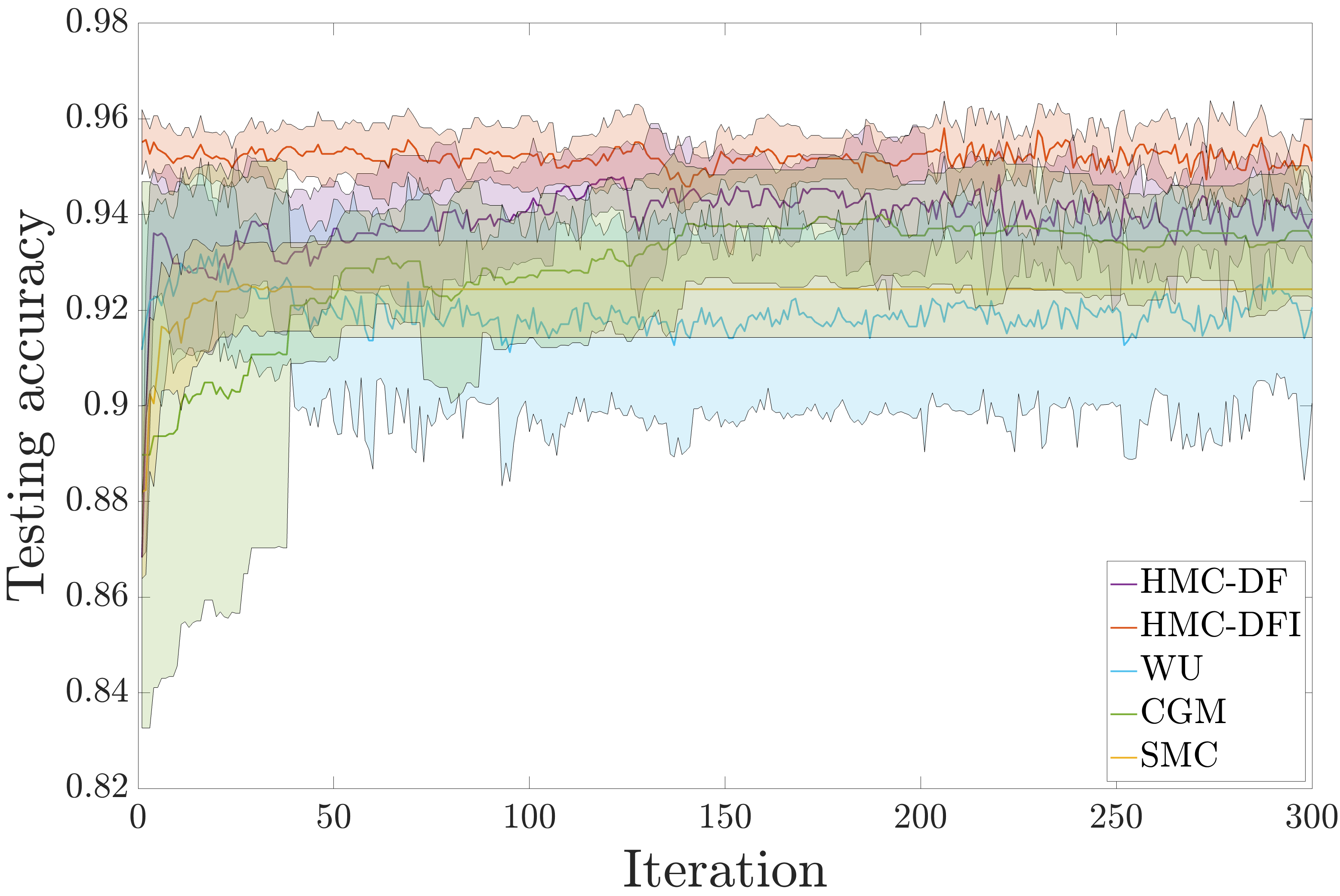}
      \caption{Breast Cancer Wisconsin dataset.}
    \end{subfigure}
    \hfill
    \begin{subfigure}[t]{.49\textwidth}
      \centering
      \includegraphics[width=\linewidth]{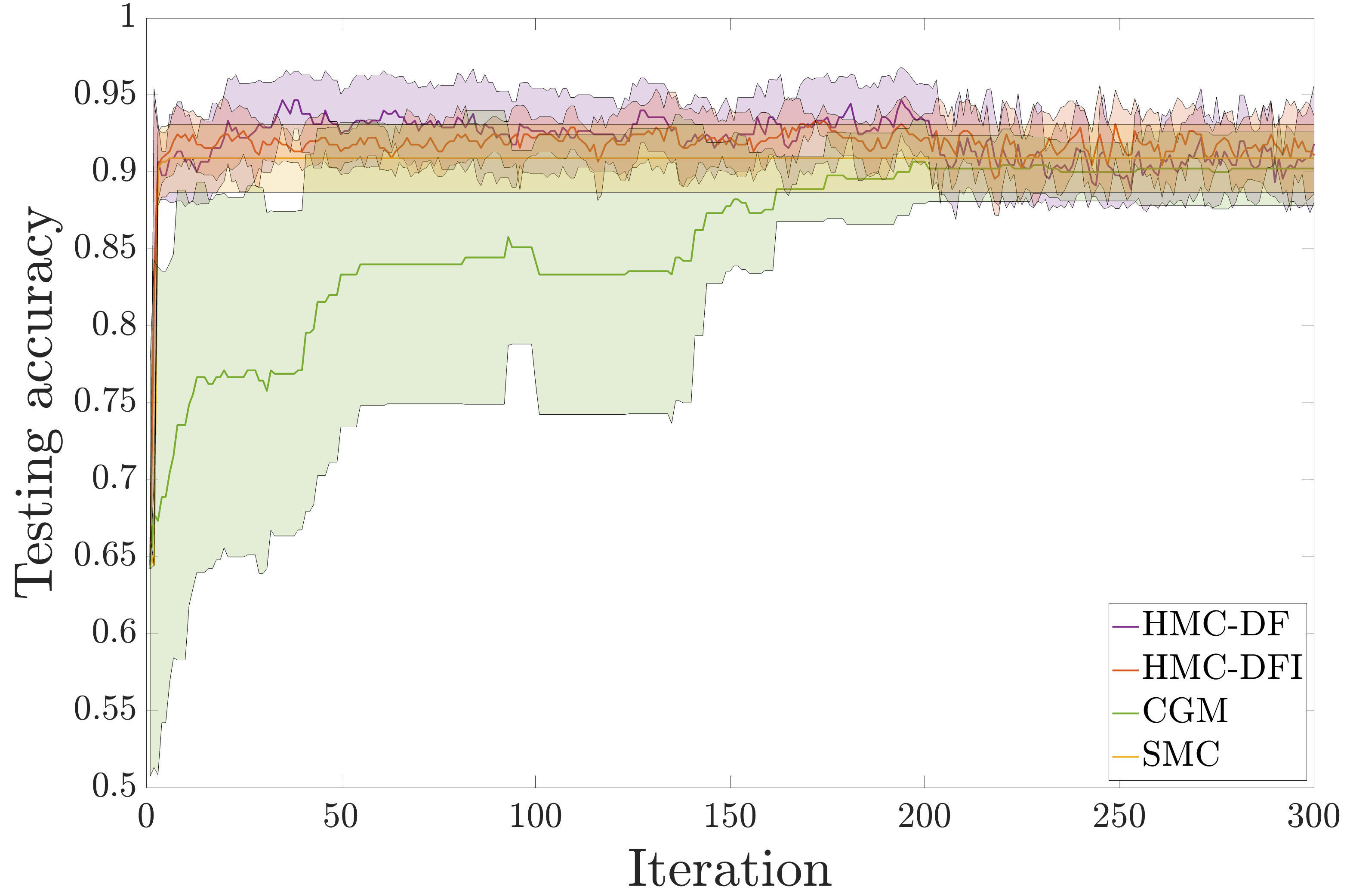}
      \caption{Iris dataset.}
    \end{subfigure}
  
    \medskip
  
    \begin{subfigure}[t]{.49\textwidth}
      \centering
      \includegraphics[width=\linewidth]{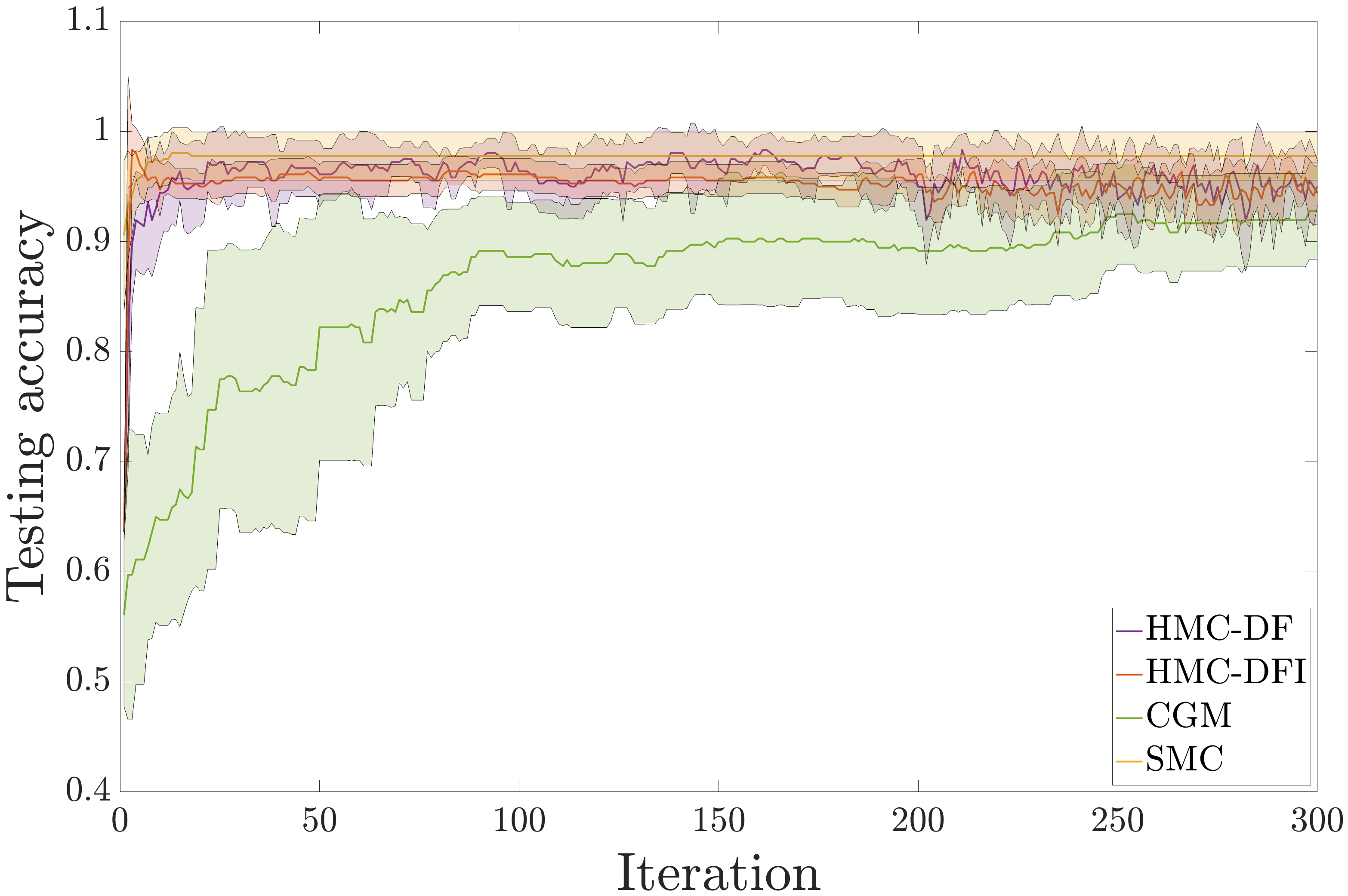}
      \caption{Wine dataset.}
    \end{subfigure}
    \hfill
    \begin{subfigure}[t]{.49\textwidth}
      \centering
      \includegraphics[width=\linewidth]{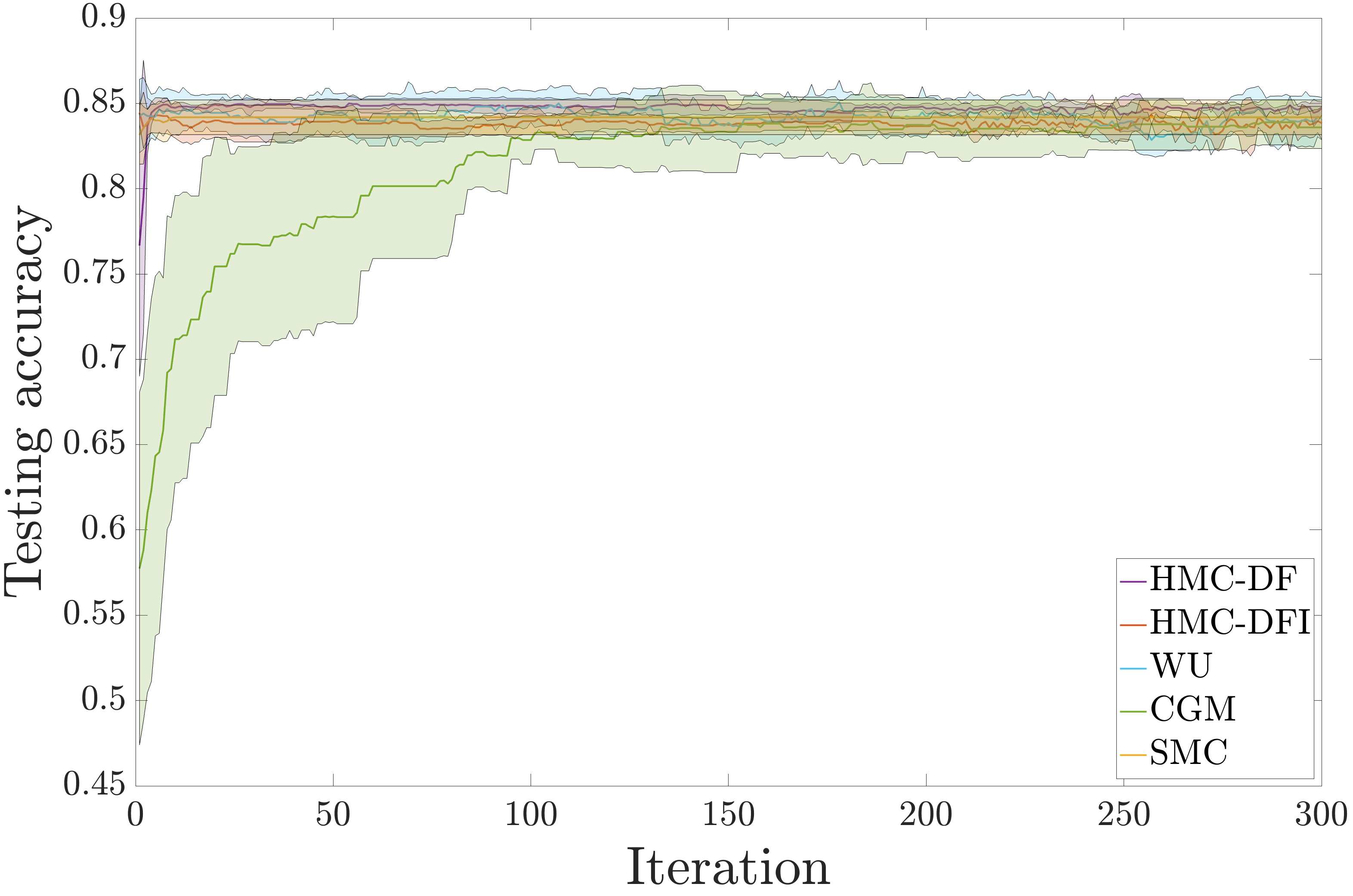}
      \caption{Raisin dataset.}
    \end{subfigure}
    \caption{Comparison of predictive accuracy on testing data between methods for real-world datasets \citep{Dua:2019}.}\label{fig:hmc-fig-metrics}
  \end{figure}

In most cases, HMC-based methods place as the top-performing method with respect to testing accuracy with improvements in overall tree complexity. Both the HMC-DF and HMC-DFI methods show significant improvement in the acceptance rate when compared to existing methods. Further visual comparison of the testing predictive accuracy across the chains produced by each method for each of the real-world datasets is shown in Figure \ref{fig:hmc-fig-metrics}. The solid line denotes the mean value across the 10 chains, and the shaded area represents one standard deviation. Note that SMC (which terminates after a criterion has been reached as opposed to a number of iterations) has been plotted such that the value given by the final set of tree particles is repeated over the rest of the chain for easy comparison.

\section{Discussion}\label{sec:discussion}
We have developed a novel method for exploring the posterior distribution of Bayesian decision trees. Likelihood information was able to be incorporated into the sampling scheme in order to jointly update the continuous parameters.  By introducing additional hyperparameters to the algorithm, we were able to implement a more efficient type of sampling method, namely HMC, which uses the geometry of the likelihood to sample the continuous tree parameters. Our approach has been implemented in two ways - where either the splitting index is discrete or continuous - and has been shown to discover trees of better testing accuracy with a higher acceptance rate, and thus movement around the posterior. 

We present both the HMC-DF and HMC-DFI approaches in this paper; although both show improvements on existing methods,  the first method is much more interpretable and therefore could be more practical to use in certain settings. In fact, the soft-input selection method moves the decision tree definition closer to that of a neural network, while still retaining some level of interpretability. Presenting each method allows either to be chosen based on the specific application of interest. 

We tested against the standard datasets in the Bayesian decision tree literature, namely, the synthetic dataset of \citet{chipman1998bayesian} and the Breast Cancer Wisconsin (Original) dataset, in addition to the Iris, Wine and Raisin datasets which are popular in the machine learning community. We feel that this gave a good range to help understand the performance of our algorithm and compare to existing methods. 

Testing on simulated data (see Table \ref{tab:synth-metrics}) showed that the HMC-based algorithms perform better than existing methods when the underlying data is generated from a tree structure. With respect to the CGM dataset, comparing the average number of leaves, and noting that the true tree has five leaf nodes, shows that our algorithm suffers less from overfitting or overexplaining than other methods. Further to this, the testing MSE is close to the expected value, given that the data was generated with a standard deviation of 0.04. Note also that when the acceptance metric of the WU multiple-proposal scheme is calculated per individual proposal, the rate is similar to the CGM method, showing that using our likelihood-based approach has helped mitigate the issue of low per-proposal acceptance rate while improving testing accuracy.

Table \ref{tab:real-metrics} shows that in most cases HMC-based methods outperform existing methods with respect to predictive testing accuracy on real-world datasets. Notably, this is achieved using a simpler tree structure, as evidenced by the low average number of leaf nodes. 
Furthermore, a significant improvement in acceptance rate is observed for both methods across all datasets. 
This is a logical result, as proposing a random value across the entire spread of the input dimension will likely result in the value being rejected if only a few proposals are viable with respect to the data.

One interesting aspect of the method is that in some cases, the biased acceptance probability leads to trees of higher likelihood, which appears to, on average, find trees with better performance metrics. This is clear when considering the results for the Iris dataset in Figure \ref{fig:hmc-fig-metrics}, where the mean and standard deviation make an observed drop at the 200 iteration mark, which aligns with the end of the burn-in period. Depending on the specific application, it may therefore be preferential to use the biased acceptance probability to find well-performing decision trees.

The optimal method for varying the value of the split sharpness parameter, $h$, is likely linked to the input characteristics. 
The final value appears to be linked to the spread in the data; larger values (and thus less sharp splits) are better when considering discrete inputs, whereas smaller values are more effective when the data is close, such as with real attributes. Further evidence of this is that the optimal final value is the same for both the HMC-DF and HMC-DFI methods for each dataset. An interesting extension of our algorithm would be to define the variation of $h$ during the adaption phase independently for each input dimension. 

There are several limitations to the methods presented in this paper. One limitation is that the tree topology update scheme still relies on the `random walk' proposal mechanism. This limitation currently exists for all Bayesian decision tree methods and has not been improved upon here. Another issue is efficiency; generating a single sample requires multiple HMC transitions, each of which is much more computationally expensive than random-walk transitions. Further to this is the need to determine suitable sampling hyperparameters (i.e. step size, mass matrix) during each intermediate transition within each iteration, which likely violates the reversibility of the chain within the practical implementation. Since the algorithm is based on a changing parameter vector, it is not clear that the sampling hyperparameters should remain the same, or how they should be adjusted, when considering a different tree model.

A logical extension of our method would be to increase the number of samples per tree topology. The current algorithm provides a single sample per tree topology - as soon as this is altered the code must be recompiled for the new model structure and the sampling hyperparameters determined via the standard NUTS approach. Further, the issue of moving to areas of high density after switching model dimensions is prevalent throughout transdimensional problems and remains here. However, incorporating HMC has clearly shown an improvement over random-walk-based methods. Therefore, investigating a way to include HMC while producing more samples per tree topology would avoid additional burn-in iterations and help further improve current performance.

\newpage
\bibliographystyle{abbrvnat}
\bibliography{main}


\newpage 
\appendix

\section{Appendix}

\subsection{General Expressions of Likelihood Derivatives}\label{sec:app-llh}

When HMC is implemented in practice, any parameter defined by a bounded distribution is transformed from the constrained space to the unconstrained space. If $q$ is taken to denote the constrained parameters and $u$ the unconstrained parameters, then the relationship is given by the constraint mapping $T:u\to q$. The log-likelihood in terms of the unconstrained parameters is given by
\begin{equation}
    \log \ell (q) = \log \ell (T(u)) := h(u).
\end{equation}
Therefore the derivative of the log-likelihood with respect to the unconstrained variable of interest $u$ is given by
\begin{equation}
    \frac{\partial}{\partial u} h(u) = \frac{\partial}{\partial T(u)} \log \ell (T(u)) \cdot \frac{\partial}{\partial u}T(u).
\end{equation}
This relationship will be used in the following derivations for those parameters that are defined by bounded distributions. Note that the variables that are defined in the unconstrained space will be denoted with the subscript ``unc''.

\subsubsection*{HMC-DF Method}

\subsubsection*{Regression}
The derivatives for the regression case can be computed using repeated application of chain rule to the logarithm of Equation~\ref{eq:llh-reg} and using Equation~\ref{eq:phi} and Equation~\ref{eq:psi}. We present general expressions for the derivatives with respect to each continuous parameter.

Both the splitting threshold $\bm{\tau}$ and the variance across leaf nodes $\sigma$ are constrained, with the constraint mappings defined as $X = (1+\exp(-Y))^{-1}$ and $X = \exp(Y)$ respectively.

\paragraph{Partial Derivative w.r.t $\bm{\tau}$}

Consider a general tree structure with $n_\ell$ number of leaf/terminal nodes. Let $\mathcal{A}(\eta_{\ell_k})$ represent the set of nodes which are ancestors of leaf node $\eta_{\ell_k}$ and $\mu_k$ the represent the corresponding mean value (assuming here that there is a constant variance $\sigma$ across all leaf nodes). Then the derivative of the log-likelihood with respect to the unconstrained splitting threshold parameter $\bm{\tau}_{unc}$ is given by,
\begin{equation}
    \frac{\partial}{\partial \tau_{j,unc}} \log\ell 
    = -\sigma^{-2} \left(\frac{\exp (-\tau_{j,unc}) }{(1+\exp(-\tau_{j,unc}))^2} \right) \sum_{i=1}^N \left(\sum_{k=1}^{n_\ell} g_k(\bm{\tau},\bm{\mu}) \right) \left( \sum_{k=1}^{n_\ell} \frac{\partial}{\partial\tau_j} g_k(\bm{\tau},\bm{\mu})\right),
\end{equation} 
where
\begin{multline}\label{eq:g_df}
    g_k(\bm{\tau},\bm{\mu}) = (\mu_{k} - y_i) \prod_{\eta \in \mathcal{A}(\eta_{\ell_k})}  \left[\left\{1+\exp\left(-\frac{\mathbf{x}_{i,\kappa_\eta}-\tau_\eta}{h}\right)\right\}^{-1}\right]^{R_{\eta}} \\ \times\left[1-\left\{1+\exp\left(-\frac{\mathbf{x}_{i,\kappa_\eta}-\tau_\eta}{h}\right)\right\}^{-1}\right]^{1-R_{\eta}},
\end{multline}
and 
\begin{multline}
    \frac{\partial}{\partial\tau_j} g_k(\bm{\tau},\bm{\mu}) = \mathbb{I}(\eta_j \in \mathcal{A}(\eta_{\ell_k}))(-1)^{R_{\eta_j}}\frac{\mu_{k}-y_i}{h}\\
    \times \exp\left(-\frac{\mathbf{x}_{i,\kappa_{\eta_j}}-\tau_j}{h}\right) \left[1+\exp\left(-\frac{\mathbf{x}_{i,\kappa_{\eta_j}}-\tau_j}{h}\right)\right]^{-2} \\
    \times \prod_{\eta \in \mathcal{A}(\eta_{\ell_k})\setminus\eta_j}  \left[\left\{1+\exp\left(-\frac{\mathbf{x}_{i,\kappa_\eta}-\tau_\eta}{h}\right)\right\}^{-1}\right]^{R_{\eta}} \left[1-\left\{1+\exp\left(-\frac{\mathbf{x}_{i,\kappa_\eta}-\tau_\eta}{h}\right)\right\}^{-1}\right]^{1-R_{\eta}}.
\end{multline}

\paragraph{Partial Derivative w.r.t $\bm{\mu}$} Considering the same general tree structure, the general expression for the derivative of the log-likelihood with respect to the leaf node mean values $\bm{\mu}$ is given by
\begin{equation}
    \frac{\partial}{\partial \mu_j} \log\ell 
    = -\sigma^{-2} \sum_{i=1}^N \left(\sum_{k=1}^{n_\ell} g_k(\bm{\tau},\bm{\mu}) \right) \left( \frac{\partial}{\partial\mu_j} g(\bm{\tau},\bm{\mu})\right),
\end{equation} 
with  $g_k(\bm{\tau},\bm{\mu})$ as defined in Equation \ref{eq:g_df} and,
\begin{multline}\label{eq:dg_dmu}
    \frac{\partial}{\partial\mu_j} g(\bm{\tau},\bm{\mu}) = \prod_{\eta \in \mathcal{A}(\eta_{\ell_j})}  \psi(\mathbf{x}_i|\mathbb{T}_\kappa,\eta)^{R_{\eta}} (1-\psi(\mathbf{x}_i|\mathbb{T}_\kappa,\eta))^{1-R_{\eta}}  \\
    = \prod_{\eta \in \mathcal{A}(\eta_{\ell_j})}  \left[\left\{1+\exp\left(-\frac{\mathbf{x}_{i,\kappa_\eta}-\tau_\eta}{h}\right)\right\}^{-1}\right]^{R_{\eta}} \left[1-\left\{1+\exp\left(-\frac{\mathbf{x}_{i,\kappa_\eta}-\tau_\eta}{h}\right)\right\}^{-1}\right]^{1-R_{\eta}}.
\end{multline}

Note the subscript in the ancestor variable in Equation \ref{eq:dg_dmu} corresponds to that of the leaf node being considered.

\paragraph{Partial Derivative w.r.t $\sigma$} Again considering the same general tree structure, the general expression for the derivative of the log-likelihood with respect to the unconstrained leaf variance $\sigma_{unc}$ is given by
\begin{align}
    \frac{\partial}{\partial\sigma_{unc}}\log\ell &= \left( -\frac{N}{\sigma} +\frac{1}{\sigma^3} \sum_{i=1}^N \left( \sum_{k=1}^{n_\ell} g_k(\bm{\tau},\bm{\mu}) \right) ^2\right) \cdot \exp(\sigma_{unc}) ,
\end{align}
 where $g_k(\bm{\tau},\bm{\mu})$ is as defined in Equation~\ref{eq:g_df}.

\subsubsection*{Classification}
Similar to the regression case, the derivatives for the classification case can be computed using repeated application of chain rule to the logarithm of Equation~\ref{eq:llh-class} and also using Equation~\ref{eq:temp}, Equation~\ref{eq:phi} and Equation~\ref{eq:psi}. We present the general expressions for the derivatives with respect to the continuous parameter $\bm{\tau}$, which again is constrained with the constraint mapping $X = (1+\exp(-Y))^{-1}$. 

\paragraph{Partial Derivative w.r.t $\bm{\tau}$}
Consider a general tree structure with $n_\ell$ number of leaf/terminal nodes. Let $\mathcal{A}(\eta_{\ell_k})$ represent the set of nodes which are ancestors of leaf node $\eta_{\ell_k}$. Then the derivative of the log-likelihood with respect to the unconstrained splitting threshold parameter $\bm{\tau}_{unc}$ is given by,
\begin{equation}
    \frac{\partial}{\partial\tau_{j,unc}}\log\ell = \left(\frac{\exp (-\tau_{j,unc}) }{(1+\exp(-\tau_{j,unc}))^2} \right)  \sum_{k=1}^{n_\ell} \left[ -\frac{\partial}{\partial\tau_j} \log (\Gamma(\Phi_k+A))+ \sum_{m=1}^M\left( \frac{\partial}{\partial\tau_j}\log(\Gamma(\bm{\phi}_{m,k}+ \alpha_m)) \right) \right],\\
\end{equation} 
where
\begin{multline}
     \frac{\partial}{\partial\tau_j} \log \Gamma(\Phi_k+A)  =  \mathbb{I}(\eta_j \in \mathcal{A}(\eta_{\ell_k}))(-1)^{R_{\eta_j}}\frac{1}{h} \Psi\left( k_j(\boldsymbol{\tau}) \right)  \\
     \times\sum_{i=1}^N  \exp\left(-\frac{\mathbf{x}_{i,\kappa_{\eta_j}}-\tau_j}{h}\right)  \left[1+\exp\left(-\frac{\mathbf{x}_{i,\kappa_{\eta_j}}-\tau_j}{h}\right)\right]^{-2} \\
    \times \prod_{\eta \in \mathcal{A}(\eta_{\ell_k})\setminus\eta_j}  \left[\left\{1+\exp\left(-\frac{\mathbf{x}_{i,\kappa_\eta}-\tau_\eta}{h}\right)\right\}^{-1}\right]^{R_{\eta}} \left[1-\left\{1+\exp\left(-\frac{\mathbf{x}_{i,\kappa_\eta}-\tau_\eta}{h}\right)\right\}^{-1}\right]^{1-R_{\eta}},
\end{multline}
\begin{multline}
    \frac{\partial}{\partial\tau_j} \log\Gamma(\bm{\phi}_{m,k}+ \alpha_m) =  \mathbb{I}(\eta_j \in \mathcal{A}(\eta_{\ell_k}))(-1)^{R_{\eta_j}}\frac{1}{h} \Psi\left( l_j(\boldsymbol{\tau}) \right)  \\
    \times\sum_{i=1}^N \mathbb{I}(y_i = m) \exp\left(-\frac{\mathbf{x}_{i,\kappa_{\eta_j}}-\tau_j}{h}\right) \left[1+\exp\left(-\frac{\mathbf{x}_{i,\kappa_{\eta_j}}-\tau_j}{h}\right)\right]^{-2} \\
    \times \prod_{\eta \in \mathcal{A}(\eta_{\ell_k})\setminus\eta_j}  \left[\left\{1+\exp\left(-\frac{\mathbf{x}_{i,\kappa_\eta}-\tau_\eta}{h}\right)\right\}^{-1}\right]^{R_{\eta}} \left[1-\left\{1+\exp\left(-\frac{\mathbf{x}_{i,\kappa_\eta}-\tau_\eta}{h}\right)\right\}^{-1}\right]^{1-R_{\eta}},
\end{multline}
where $\Psi(\cdot)$ is the digamma function, and 
\begin{align}
    k_j(\boldsymbol{\tau}) &= A + \sum_{i=1}^N \left[ \prod_{\eta \in \mathcal{A}(\eta_{\ell_k})}  \left(\left\{1+\exp\left(-\frac{\mathbf{x}_{i,\kappa_\eta}-\tau_\eta}{h}\right)\right\}^{-1}\right)^{R_{\eta}}\right.\nonumber\\
    &\quad\quad\quad\quad\quad\quad\quad\quad\quad\quad\times\left.\left(1-\left\{1+\exp\left(-\frac{\mathbf{x}_{i,\kappa_\eta}-\tau_\eta}{h}\right)\right\}^{-1}\right)^{1-R_{\eta}} \right],\\
    l_j(\boldsymbol{\tau}) &= \alpha_m +  \sum_{i=1}^N \left[ \mathbb{I}(y_i = m)\prod_{\eta \in \mathcal{A}(\eta_{\ell_k})}  \left(\left\{1+\exp\left(-\frac{\mathbf{x}_{i,\kappa_\eta}-\tau_\eta}{h}\right)\right\}^{-1}\right)^{R_{\eta}}\right. \nonumber\\
    &\quad\quad\quad\quad\quad\quad\quad\quad\quad\quad\times\left.\left(1-\left\{1+\exp\left(-\frac{\mathbf{x}_{i,\kappa_\eta}-\tau_\eta}{h}\right)\right\}^{-1}\right)^{1-R_{\eta}} \right].
\end{align}

\subsubsection*{HMC-DFI Method}

\subsubsection*{Regression}
Similar to before, the derivatives for the regression case can be computed using repeated application of chain rule to the logarithm of Equation~\ref{eq:llh-reg} but now using Equation~\ref{eq:phi} and Equation~\ref{eq:psi-index}. We present the general expressions for the derivatives with respect to each parameter, which in this case are all continuous.

The splitting threshold $\bm{\tau}$ and the variance across leaf nodes $\sigma$ are constrained, with the constraint mappings defined as $X = (1+\exp(-Y))^{-1}$ and $X = \exp(Y)$ respectively. The splitting simplex $\bm{\Delta}$ variable is also constrained to the unit simplex, with constraint mapping defined by the stick-breaking process. First, intermediate variables that represent break proportion ratios are computed via,
\begin{equation*}
    z_k = \text{logit}^{-1}\left(y_k+\log\left( \frac{1}{K-k} \right) \right),
\end{equation*}
which then defines the ``amount'' of the unit interval that is assigned to each transformed variable,
\begin{align*}
    x_k &= \left( 1-\sum_{k'=1}^{k-1} \right) z_k, \quad k=1,\dots,K-1 \\
    x_K &= 1 - \sum_{k=1}^{K-1} x_k
\end{align*}
The derivative of this constraint transformation is defined recursively, refer to \textsc{Stan} Reference Manual Section 10.7 \cite{stan2019} for information on how to compute this value. This derivative term will be denoted by $\frac{\partial T}{\partial \Delta_{unc}}$ in the following equations.

\paragraph{Partial Derivative w.r.t $\bm{\tau}$}

Consider a general tree structure with $n_\ell$ number of leaf/terminal nodes. Let $\mathcal{A}(\eta_{\ell_k})$ represent the set of nodes which are ancestors of leaf node $\eta_{\ell_k}$ and $\mu_k$ the represent the corresponding mean value (assuming here that there is a constant variance $\sigma$ across all leaf nodes). Then the derivative of the log-likelihood with respect to the unconstrained splitting threshold parameter $\bm{\tau}_{unc}$ is given by,
\begin{equation}
    \frac{\partial}{\partial \tau_{j,unc}} \log\ell 
    = -\sigma^{-2} \left(\frac{\exp (-\tau_{j,unc}) }{(1+\exp(-\tau_{j,unc}))^2} \right) \sum_{i=1}^N \left(\sum_{k=1}^{n_\ell} g_k(\bm{\tau},\bm{\mu}) \right) \left( \sum_{k=1}^{n_\ell} \frac{\partial}{\partial\tau_j} g_k(\bm{\tau},\bm{\mu})\right),
\end{equation} 
where
\begin{multline}\label{eq:g-dfi}
    g_k(\bm{\tau},\bm{\mu}) = (\mu_{k} - y_i) \prod_{\eta \in \mathcal{A}(\eta_{\ell_k})}  \left[\left\{1+\exp\left(-\frac{\mathbf{x}_i\Delta_\eta-\tau_\eta}{h}\right)\right\}^{-1}\right]^{R_{\eta}} \\ \times\left[1-\left\{1+\exp\left(-\frac{\mathbf{x}_i\Delta_\eta-\tau_\eta}{h}\right)\right\}^{-1}\right]^{1-R_{\eta}},
\end{multline}
and 
\begin{multline}
    \frac{\partial}{\partial\tau_j} g_k(\bm{\tau},\bm{\mu}) = \mathbb{I}(\eta_j \in \mathcal{A}(\eta_{\ell_k}))(-1)^{R_{\eta_j}}\frac{\mu_{k}-y_i}{h}\\
    \times \exp\left(-\frac{\mathbf{x}_i\Delta_{\eta_j}-\tau_j}{h}\right) \left[1+\exp\left(-\frac{\mathbf{x}_i\Delta_{\eta_j}-\tau_j}{h}\right)\right]^{-2} \\
    \times \prod_{\eta \in \mathcal{A}(\eta_{\ell_k})\setminus\eta_j}  \left[\left\{1+\exp\left(-\frac{\mathbf{x}_i\Delta_\eta-\tau_\eta}{h}\right)\right\}^{-1}\right]^{R_{\eta}} \left[1-\left\{1+\exp\left(-\frac{\mathbf{x}_i\Delta_\eta-\tau_\eta}{h}\right)\right\}^{-1}\right]^{1-R_{\eta}}.
\end{multline}

\paragraph{Partial Derivative w.r.t $\bm{\Delta}$} Considering the same general tree structure, the general expression for the derivative of the log-likelihood with respect to the splitting simplexes $\bm{\Delta}$ is given by

\begin{equation}
    \frac{\partial}{\partial \Delta_{j,unc}} \log\ell 
    = -\sigma^{-2} \frac{\partial T}{\partial \Delta_{j,unc}} \sum_{i=1}^N \left(\sum_{k=1}^{n_\ell} g_k(\bm{\tau},\bm{\mu}) \right) \left( \sum_{k=1}^{n_\ell} \frac{\partial}{\partial\Delta_j} g_k(\bm{\tau},\bm{\mu})\right),
\end{equation} 
where $g_k(\bm{\tau},\bm{\mu}) $ is as defined above in Equation \ref{eq:g-dfi} and,
\begin{multline}
    \frac{\partial}{\partial\Delta_j} g_k(\bm{\tau},\bm{\mu}) = \mathbb{I}(\eta_j \in \mathcal{A}(\eta_{\ell_k}))(-1)^{(1-R_{\eta_j})}\frac{(\mu_{k}-y_i)\cdot \mathbf{x}_i}{h}\\
    \times \exp\left(-\frac{\mathbf{x}_i\Delta_{\eta_j}-\tau_j}{h}\right) \left[1+\exp\left(-\frac{\mathbf{x}_i\Delta_{\eta_j}-\tau_j}{h}\right)\right]^{-2} \\
    \times \prod_{\eta \in \mathcal{A}(\eta_{\ell_k})\setminus\eta_j}  \left[\left\{1+\exp\left(-\frac{\mathbf{x}_i\Delta_\eta-\tau_\eta}{h}\right)\right\}^{-1}\right]^{R_{\eta}} \left[1-\left\{1+\exp\left(-\frac{\mathbf{x}_i\Delta_\eta-\tau_\eta}{h}\right)\right\}^{-1}\right]^{1-R_{\eta}}.
\end{multline}

\paragraph{Partial Derivative w.r.t $\bm{\mu}$} Considering the same general tree structure, the general expression for the derivative of the log-likelihood with respect to the leaf node mean values $\bm{\mu}$ is given by
\begin{equation}
    \frac{\partial}{\partial \mu_j} \log\ell 
    = -\sigma^{-2} \sum_{i=1}^N \left(\sum_{k=1}^{n_\ell} g_k(\bm{\tau},\bm{\mu}) \right) \left( \frac{\partial}{\partial\mu_j} g(\bm{\tau},\bm{\mu})\right),
\end{equation} 
with $g_k(\bm{\tau},\bm{\mu}) $ as defined above in Equation \ref{eq:g-dfi} and,
\begin{multline}\label{eq:dg_dmu-dfi}
    \frac{\partial}{\partial\mu_j} g(\bm{\tau},\bm{\mu}) = \prod_{\eta \in \mathcal{A}(\eta_{\ell_j})}  \psi(\mathbf{x}_i|\mathbb{T}_\Delta,\eta)^{R_{\eta}} (1-\psi(\mathbf{x}_i|\mathbb{T}_\Delta,\eta))^{1-R_{\eta}} \\
    = \prod_{\eta \in \mathcal{A}(\eta_{\ell_j})}  \left[\left\{1+\exp\left(-\frac{\mathbf{x}_i\Delta_\eta-\tau_\eta}{h}\right)\right\}^{-1}\right]^{R_{\eta}} \left[1-\left\{1+\exp\left(-\frac{\mathbf{x}_i\Delta_\eta-\tau_\eta}{h}\right)\right\}^{-1}\right]^{1-R_{\eta}}.
\end{multline}

Note the subscript in the ancestor variable in Equation~\ref{eq:dg_dmu-dfi} corresponds to that of the leaf node being considered.

\paragraph{Partial Derivative w.r.t $\sigma$} Again considering the same general tree structure, the general expression for the derivative of the log-likelihood with respect to the unconstrained leaf variance $\sigma_{unc}$ is given by
\begin{align}
    \frac{\partial}{\partial\sigma_{unc}}\log\ell &= \left( -\frac{N}{\sigma} +\frac{1}{\sigma^3} \sum_{i=1}^N \left( \sum_{k=1}^{n_\ell} g_k(\bm{\tau},\bm{\mu}) \right) ^2\right) \cdot \exp(\sigma_{unc}) ,
\end{align}
 where $g_k(\bm{\tau},\bm{\mu})$ is as defined in Equation~\ref{eq:g-dfi}.

\subsubsection*{Classification}
Similar to the regression case, the derivatives for the classification case can be computed using repeated application of chain rule to the logarithm of Equation~\ref{eq:llh-class} and also using Equation~\ref{eq:temp}, Equation~\ref{eq:phi} and Equation~\ref{eq:psi-index}. The general expressions for the derivatives with respect to the continuous parameter $\bm{\tau}$ are presented here.

\paragraph{Partial Derivative w.r.t $\bm{\tau}$} 
Consider a general tree structure with $n_\ell$ number of leaf/terminal nodes. Let $\mathcal{A}(\eta_{\ell_k})$ represent the set of nodes which are ancestors of leaf node $\eta_{\ell_k}$. Then the derivative of the log-likelihood with respect to the unconstrained splitting threshold parameter $\bm{\tau}_{unc}$ is given by,
\begin{equation}
    \frac{\partial}{\partial\tau_{j,unc}}\log\ell = \left(\frac{\exp (-\tau_{j,unc}) }{(1+\exp(-\tau_{j,unc}))^2} \right)  \sum_{k=1}^{n_\ell} \left[ -\frac{\partial}{\partial\tau_j} \log (\Gamma(\Phi_k+A))+ \sum_{m=1}^M\left( \frac{\partial}{\partial\tau_j}\log(\Gamma(\bm{\phi}_{m,k}+ \alpha_m)) \right) \right],\\
\end{equation} 
where
\begin{multline}
     \frac{\partial}{\partial\tau_j} \log \Gamma(\Phi_k+A)  =  \mathbb{I}(\eta_j \in \mathcal{A}(\eta_{\ell_k}))(-1)^{R_{\eta_j}}\frac{1}{h} \Psi\left( k_j(\boldsymbol{\tau}) \right)  \\
     \times\sum_{i=1}^N  \exp\left(-\frac{\mathbf{x}_i\Delta_{\eta_j}-\tau_j}{h}\right)  \left[1+\exp\left(-\frac{\mathbf{x}_i\Delta_{\eta_j}-\tau_j}{h}\right)\right]^{-2} \\
    \times \prod_{\eta \in \mathcal{A}(\eta_{\ell_k})\setminus\eta_j}  \left[\left\{1+\exp\left(-\frac{\mathbf{x}_i\Delta_\eta-\tau_\eta}{h}\right)\right\}^{-1}\right]^{R_{\eta}} \left[1-\left\{1+\exp\left(-\frac{\mathbf{x}_i\Delta_\eta-\tau_\eta}{h}\right)\right\}^{-1}\right]^{1-R_{\eta}},
\end{multline}
\begin{multline}
    \frac{\partial}{\partial\tau_j} \log\Gamma(\bm{\phi}_{m,k}+ \alpha_m) =  \mathbb{I}(\eta_j \in \mathcal{A}(\eta_{\ell_k}))(-1)^{R_{\eta_j}}\frac{1}{h} \Psi\left( l_j(\boldsymbol{\tau}) \right)  \\
    \times\sum_{i=1}^N \mathbb{I}(y_i = m) \exp\left(-\frac{\mathbf{x}_i\Delta_{\eta_j}-\tau_j}{h}\right) \left[1+\exp\left(-\frac{\mathbf{x}_i\Delta_{\eta_j}-\tau_j}{h}\right)\right]^{-2} \\
    \times \prod_{\eta \in \mathcal{A}(\eta_{\ell_k})\setminus\eta_j}  \left[\left\{1+\exp\left(-\frac{\mathbf{x}_i\Delta_\eta-\tau_\eta}{h}\right)\right\}^{-1}\right]^{R_{\eta}} \left[1-\left\{1+\exp\left(-\frac{\mathbf{x}_i\Delta_\eta-\tau_\eta}{h}\right)\right\}^{-1}\right]^{1-R_{\eta}},
\end{multline}
where $\Psi(\cdot)$ is the digamma function, and 
\begin{align}
    k_j(\boldsymbol{\tau}) &= A + \sum_{i=1}^N \left[ \prod_{\eta \in \mathcal{A}(\eta_{\ell_k})}  \left(\left\{1+\exp\left(-\frac{\mathbf{x}_i\Delta_\eta-\tau_\eta}{h}\right)\right\}^{-1}\right)^{R_{\eta}}\right.\nonumber\\
    &\quad\quad\quad\quad\quad\quad\quad\quad\quad\quad\times\left.\left(1-\left\{1+\exp\left(-\frac{\mathbf{x}_i\Delta_\eta-\tau_\eta}{h}\right)\right\}^{-1}\right)^{1-R_{\eta}} \right],\\
    l_j(\boldsymbol{\tau}) &= \alpha_m +  \sum_{i=1}^N \left[ \mathbb{I}(y_i = m)\prod_{\eta \in \mathcal{A}(\eta_{\ell_k})}  \left(\left\{1+\exp\left(-\frac{\mathbf{x}_i\Delta_\eta-\tau_\eta}{h}\right)\right\}^{-1}\right)^{R_{\eta}}\right. \nonumber\\
    &\quad\quad\quad\quad\quad\quad\quad\quad\quad\quad\times\left.\left(1-\left\{1+\exp\left(-\frac{\mathbf{x}_i\Delta_\eta-\tau_\eta}{h}\right)\right\}^{-1}\right)^{1-R_{\eta}} \right].
\end{align}

\paragraph{Partial Derivative w.r.t $\bm{\Delta}$} Considering the same general tree structure, the general expression for the derivative of the log-likelihood with respect to the splitting simplexes $\bm{\Delta}$ is given by

\begin{equation}
    \frac{\partial}{\partial\Delta_{j,unc}}\log\ell = \frac{\partial T}{\partial \Delta_{j,unc}} \sum_{k=1}^{n_\ell} \left[ -\frac{\partial}{\partial\Delta_j} \log (\Gamma(\Phi_k+A))+ \sum_{m=1}^M\left( \frac{\partial}{\partial\Delta_j}\log(\Gamma(\bm{\phi}_{m,k}+ \alpha_m)) \right) \right],\\
\end{equation} 
where
\begin{multline}
     \frac{\partial}{\partial\Delta_j} \log \Gamma(\Phi_k+A)  =  \mathbb{I}(\eta_j \in \mathcal{A}(\eta_{\ell_k}))(-1)^{(1-R_{\eta_j})}\Psi\left( k_j(\boldsymbol{\tau}) \right)  \\
     \times\sum_{i=1}^N \frac{\mathbf{x}_i}{h}  \exp\left(-\frac{\mathbf{x}_i\Delta_{\eta_j}-\tau_j}{h}\right)  \left[1+\exp\left(-\frac{\mathbf{x}_i\Delta_{\eta_j}-\tau_j}{h}\right)\right]^{-2} \\
    \times \prod_{\eta \in \mathcal{A}(\eta_{\ell_k})\setminus\eta_j}  \left[\left\{1+\exp\left(-\frac{\mathbf{x}_i\Delta_\eta-\tau_\eta}{h}\right)\right\}^{-1}\right]^{R_{\eta}} \left[1-\left\{1+\exp\left(-\frac{\mathbf{x}_i\Delta_\eta-\tau_\eta}{h}\right)\right\}^{-1}\right]^{1-R_{\eta}},
\end{multline}
\begin{multline}
    \frac{\partial}{\partial\Delta_j} \log\Gamma(\bm{\phi}_{m,k}+ \alpha_m) =  \mathbb{I}(\eta_j \in \mathcal{A}(\eta_{\ell_k}))(-1)^{(1-R_{\eta_j})} \Psi\left( l_j(\boldsymbol{\tau}) \right)  \\
    \times\sum_{i=1}^N  \mathbb{I}(y_i = m) \frac{\mathbf{x}_i}{h} \exp\left(-\frac{\mathbf{x}_i\Delta_{\eta_j}-\tau_j}{h}\right) \left[1+\exp\left(-\frac{\mathbf{x}_i\Delta_{\eta_j}-\tau_j}{h}\right)\right]^{-2} \\
    \times \prod_{\eta \in \mathcal{A}(\eta_{\ell_k})\setminus\eta_j}  \left[\left\{1+\exp\left(-\frac{\mathbf{x}_i\Delta_\eta-\tau_\eta}{h}\right)\right\}^{-1}\right]^{R_{\eta}} \left[1-\left\{1+\exp\left(-\frac{\mathbf{x}_i\Delta_\eta-\tau_\eta}{h}\right)\right\}^{-1}\right]^{1-R_{\eta}},
\end{multline}
where $\Psi(\cdot)$ is the digamma function, and 
\begin{align}
    k_j(\boldsymbol{\tau}) &= A + \sum_{i=1}^N \left[ \prod_{\eta \in \mathcal{A}(\eta_{\ell_k})}  \left(\left\{1+\exp\left(-\frac{\mathbf{x}_i\Delta_\eta-\tau_\eta}{h}\right)\right\}^{-1}\right)^{R_{\eta}}\right.\nonumber\\
    &\quad\quad\quad\quad\quad\quad\quad\quad\quad\quad\times\left.\left(1-\left\{1+\exp\left(-\frac{\mathbf{x}_i\Delta_\eta-\tau_\eta}{h}\right)\right\}^{-1}\right)^{1-R_{\eta}} \right],\\
    l_j(\boldsymbol{\tau}) &= \alpha_m +  \sum_{i=1}^N \left[ \mathbb{I}(y_i = m)\prod_{\eta \in \mathcal{A}(\eta_{\ell_k})}  \left(\left\{1+\exp\left(-\frac{\mathbf{x}_i\Delta_\eta-\tau_\eta}{h}\right)\right\}^{-1}\right)^{R_{\eta}}\right. \nonumber\\
    &\quad\quad\quad\quad\quad\quad\quad\quad\quad\quad\times\left.\left(1-\left\{1+\exp\left(-\frac{\mathbf{x}_i\Delta_\eta-\tau_\eta}{h}\right)\right\}^{-1}\right)^{1-R_{\eta}} \right].
\end{align}

\newpage
\subsection{Derivation of Acceptance Probability} \label{sec:rjhmc-tree-move}

This section will first discuss the grow and prune moves and the associated proposal ratio required to adhere to the RJMCMC transition. This method of moving within the transdimensional model is based on the RJMCMC method as originally presented in \citet{green1995reversible}. Note that the stay proposal does not change the tree topology $\mathcal{T}$ and therefore does not require transdimensional consideration. The transdimensional proposal ratio therefore simplifies to one. Following this, the derivation of the double-sided RJHMC acceptance probability will be presented.

The discussion presented here has been defined with respect to the HMC-DF definition, noting that a simple substitution of $\kappa$ to $\Delta$ gives the correct equations for the model used for the HMC-DFI definition. As the expressions apply to both RJHMC-Tree methods, the likelihood and prior terms are left general. See Sections \ref{sec:likelihoods} and \ref{sec:prior} for the relevant definitions.

\subsubsection*{Grow Proposal}

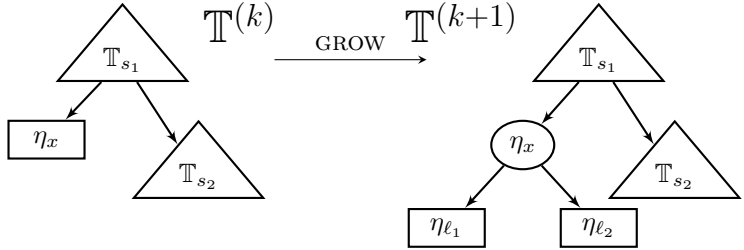
\begin{figure}[ht!]
    \begin{center}
        \begin{tikzpicture}
            \node [subtree]                                 (r1)    {$\T_{s_1}$};
            \node [subtree, right=5cm of r1]                (r2)    {$\T_{s_1}$};
            \node [subtree, below=0.5cm of r1, xshift=1cm]  (t1)    {$\T_{s_2}$};
            \node [subtree, below=0.5cm of r2, xshift=1cm]  (t2)    {$\T_{s_2}$};
            \node [leaf, below=0.5cm of r1, xshift=-1cm]    (n1)    {$\eta_x$};
            \node [internal, below=0.5cm of r2, xshift=-1cm](n2)    {$\eta_x$};
            \node [leaf, below=0.5cm of n2, xshift=-1cm]    (l1)    {$\eta_{\ell_1}$};
            \node [leaf, below=0.5cm of n2, xshift=1cm]     (l2)    {$\eta_{\ell_2}$};
            \node[right=0.3cm of r1, yshift=0.5cm] {\LARGE$\T^{(k)}$};
            \node[left=0.3cm of r2, yshift=0.5cm] {\LARGE$\T^{(k+1)}$};
            \path [line] (r1) -- (n1);
            \path [line] (r2) -- (n2);
            \path [line] (r1) -- (t1);
            \path [line] (r2) -- (t2);
            \path [line] (n2) -- (l1);
            \path [line] (n2) -- (l2);
            \draw [->,>=stealth] (2,0) -- node[above] {\textsc{grow}} (4,0);
        \end{tikzpicture}
        \caption{Grow transition from tree $\T^{(k)}$ with $n_\ell = k$ leaf nodes to tree $\T^{(k+1)}$ with $n_\ell = k+1$ leaf nodes by a using grow proposal at $\eta_x$. $\T_{s_1}$ and $\T_{s_2}$ denote generic subtrees.}
        \label{fig:grow}
    \end{center}
\end{figure}

A grow proposal is defined to be a move that increases the model dimension. Consider the transition from tree $\T^{(k)}$ to $\T^{(k+1)}$ as shown in Figure \ref{fig:grow}. If the number of leaf nodes is denoted $n_\ell = k$, it can be seen that under this proposal, the model dimension has increased by one, i.e. $k\to k+1$. 

An additional internal node and leaf node have been introduced in the new decision tree. To ensure the dimension-matching criterion is satisfied, the state vector of the initial tree is augmented with additional random variables (with appropriate change of indexing if required), to give the grow transformation as follows,
\begin{equation}\label{eq:proof-grow}
    \left( \begin{matrix} \tau_1^{(k)} \\ \vdots \\  \tau_k^{(k)} \\ u_1^{(k)} \\ \kappa_1^{(k)} \\ \vdots \\ \kappa_k^{(k)}  \\ u_2^{(k)} \\ \theta_1^{(k)} \\ \vdots \\ \theta_{k+1}^{(k)} \\ u_3^{(k)} \end{matrix} \right) \xrightarrow{T_{\textsc{grow}}(\cdot)}  \left( \begin{matrix} \tau_1^{(k+1)} \\ \vdots \\  \tau_k^{(k+1)} \\ \tau_{k+1}^{(k+1)}  \\ \kappa_1^{(k+1)} \\ \vdots \\ \kappa_k^{(k+1)}  \\ \kappa_{k+1}^{(k+1)}  \\ \theta_1^{(k+1)} \\ \vdots \\ \theta_{k+1}^{(k+1)} \\ \theta_{k+2}^{(k+1)} \end{matrix} \right).
\end{equation}
This example shows the case where a grow move has been applied to a regression tree, which requires three random variables $u_1$, $u_2$ and $u_3$ (here, $u$ is used to remain consistent with the original notation presented in \citet{green1995reversible}). In the case of classification trees, only the first two are required. The additional random variables are drawn from their respective priors as defined in Section \ref{sec:prior}. 
Note that this also holds true for the HMC-DFI definition with a simple change of notation. 

The probability of selecting to transition from $\T^i$ to $\T^*$ via a grow proposal is computed as the probability of choosing the grow move multiplied by the probability of choosing a node at which to grow. Assuming that there are $k$ leaves in $\T^i$, the probability of transitioning from tree $\T^i$ to $\T^*$ via a grow move is calculated as:
\begin{equation*}
    q_{mn}\left( \T^*  \mid  \T^i \right) = p_{\textsc{grow}}  \times \frac{1}{k} \\
\end{equation*}
The calculation for the reverse step requires computing the number of ways to decrease the model dimension by one when moving from $\T^*$ to $\T^i$. Note that this is not directly related to the number of leaf nodes present, as some proposals would result in a reduction of multiple dimensions. To account for this, only internal nodes for which both children are leaf nodes are considered for the prune move. If this number is denoted $n_f$, the probability of proposing the reverse move is then calculated as:
\begin{equation*}
    q_{nm}\left( \T^i  \mid  \T^* \right) = p_{\textsc{prune}}  \times \frac{1}{n_f} \\
\end{equation*}
To account for the dimension-matching criterion due to the change in dimension, the new random variables are drawn from their respective priors as follows,
\begin{equation*}
    u_1 \sim p_\tau(\tau \mid \mathcal{T}),\quad u_2 \sim p_\kappa(\kappa \mid \mathcal{T}), \quad u_3 \sim p_\theta(\theta \mid \mathcal{T}).
\end{equation*}
This results in the proposal ratio of,
\begin{equation}
    \frac{q_{nm}\left( \T^i  \mid  \T^* \right)}{ q_{mn}\left( \T^*  \mid  \T^i \right)} = \frac{p_{\textsc{prune}}   \frac{1}{n_f}  }{p_{\textsc{grow}}   \frac{1}{k}  p_\tau(u_1 \mid \mathcal{T})p_\kappa(u_2 \mid \mathcal{T})p_\theta(u_3 \mid \mathcal{T})}.
\end{equation}

\subsubsection*{Prune Proposal}

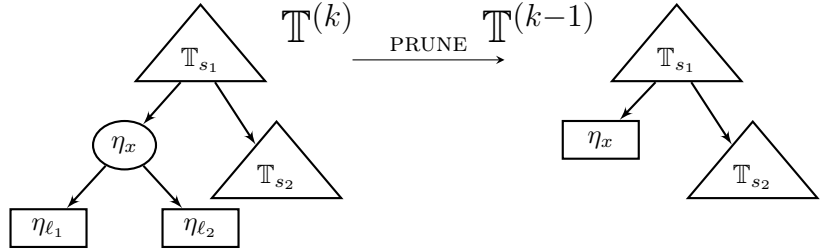
\begin{figure}[ht!]
    \begin{center}
        \begin{tikzpicture}
            \node [subtree]                                 (r1)    {$\T_{s_1}$};
            \node [subtree, right=5cm of r1]                (r2)    {$\T_{s_1}$};
            \node [subtree, below=0.5cm of r1, xshift=1cm]  (t1)    {$\T_{s_2}$};
            \node [subtree, below=0.5cm of r2, xshift=1cm]  (t2)    {$\T_{s_2}$};
            \node [internal, below=0.5cm of r1, xshift=-1cm](n1)    {$\eta_x$};
            \node [leaf, below=0.5cm of n1, xshift=-1cm]    (l1)    {$\eta_{\ell_1}$};
            \node [leaf, below=0.5cm of n1, xshift=1cm]     (l2)    {$\eta_{\ell_2}$};
            \node [leaf, below=0.5cm of r2, xshift=-1cm]    (n2)    {$\eta_x$};
            \node[right=0.3cm of r1, yshift=0.5cm] {\LARGE$\T^{(k)}$};
            \node[left=0.3cm of r2, yshift=0.5cm] {\LARGE$\T^{(k-1)}$};
            \path [line] (r1) -- (n1);
            \path [line] (r2) -- (n2);
            \path [line] (r1) -- (t1);
            \path [line] (r2) -- (t2);
            \path [line] (n1) -- (l1);
            \path [line] (n1) -- (l2);
            \draw [->,>=stealth] (2,0) -- node[above] {\textsc{prune}} (4,0);
        \end{tikzpicture}
        \caption{Transition from $\T^{(k)}$ with $n_\ell=k$ leaf nodes to $\T^{(k-1)}$ with $n_\ell=k-1$ leaf nodes via a prune proposal at $\eta_x$. $\T_{s_1}$ and $\T_{s_2}$ denote generic subtrees.}
        \label{fig:prune}
    \end{center}
\end{figure}

Consider the prune move in which a move is proposed that reduces the model dimension by one, as shown in  Figure \ref{fig:prune} moving from tree $\T^{(k)}$ to $\T^{(k-1)}$.

In this case, the state vector of the proposed tree must be augmented with additional random variables to uphold the dimension-matching condition. The extra variables represent those that would need to be drawn to compute the reverse move. The prune transformation mapping for a regression tree can be expressed as follows,

\begin{equation}\label{eq:proof-prune}
    \left( \begin{matrix} \tau_1^{(k)} \\ \vdots \\  \tau_{k-1}^{(k)}  \\  \tau_k^{(k)}  \\ \kappa_1^{(k)} \\ \vdots \\ \kappa_{k-1}^{(k)} \\ \kappa_k^{(k)}  \\ \theta_1^{(k)} \\ \vdots \\ \theta_{k}^{(k)}  \\ \theta_{k+1}^{(k)}  \end{matrix} \right) \xrightarrow{T_{\textsc{prune}}(\cdot) }  \left( \begin{matrix} \tau_1^{(k-1)} \\ \vdots \\  \tau_{k-1}^{(k-1)} \\ u_1^{(k-1)} \\  \kappa_1^{(k-1)} \\ \vdots \\ \kappa_{k-1}^{(k-1)} \\ u_2^{(k-1)}  \\ \theta_1^{(k-1)} \\ \vdots \\ \theta_{k}^{(k-1)} \\ u_3^{(k-1)} \end{matrix} \right) 
\end{equation}
where again the additional random variables are drawn from their respective priors. The transformation is equivalent for classification trees but where the $\theta$ variables and associated random variables are not required.

In a similar manner to before, the probability of a prune transition is the probability of selecting the prune move multiplied by the probability of selecting a valid node at which to prune. To calculate the probability of transitioning from tree $\T^i$ to $\T^*$ via the prune proposal, the number of internal nodes for which both children nodes are leaves need to be determined (this is again denoted by $n_f$). The probability is then calculated as,
\begin{equation*}
    q_{mn}\left( \T^*  \mid  \T^i \right) = p_{\textsc{prune}}  \times \frac{1}{n_f} \\
\end{equation*}
Assuming again that there are $k$ leaf nodes in tree $\T^i$, the probability of the reverse move from tree $\T^*$ to tree $\T^i$ is given by,
\begin{equation*}
    q_{nm}\left( \T^i  \mid  \T^* \right) = p_{\textsc{grow}}  \times \frac{1}{k-1} \\
\end{equation*}
Again, the new random variables required for the dimension-matching criterion are drawn from their respective priors as follows,
\begin{equation*}
    u_1 \sim p_\tau(\tau \mid \mathcal{T}),\quad u_2 \sim p_\kappa(\kappa \mid \mathcal{T}), \quad u_3 \sim p_\theta(\theta \mid \mathcal{T}).
\end{equation*}
The proposal ratio for the acceptance probability is therefore given by,
\begin{equation}
    \frac{q_{nm}\left( \T^i  \mid  \T^* \right)}{q_{mn}\left( \T^*  \mid  \T^i \right)}= \frac{p_{\textsc{grow}}   \frac{1}{k-1}  p_\tau(u_1 \mid \mathcal{T})p_\kappa(u_2 \mid \mathcal{T})p_\theta(u_3 \mid \mathcal{T})}{p_{\textsc{prune}}   \frac{1}{n_f}  }.
\end{equation}

\subsubsection*{Double-Sided RJHMC Acceptance Probability}

It is logical to extend the RJHMC algorithm such that movement is taken on both sides of the transdimensional step, not just on the higher dimension. The original improvement to the RJMCMC proposed in \citet{al2004improving} derived the single-sided expression, which was adopted in the RJHMC algorithm \cite{sen2017transdimensional}. This section rederives the acceptance probability for the double-sided case.

Keeping with the notation used in the original paper, let $m$ denote the initial dimension and $n$ denote the final dimension of the transition. Further, let the current sample be denoted $\mathbf{x}$ and the proposed sample $\mathbf{x}^*$. Denote the intermediate proposal distributions in the initial dimension and final dimensions as $\pi_m^*$ and $\pi_n^*$ respectively. Note that both intermediate distribution transitions correspond to movement through the augmented target distribution used within the HMC method. 

The forward transition involves first moving to a new proposal $\mathbf{x}'$ in the same dimension via $k$ HMC steps with respect to the intermediate distribution $\pi_m^*$, changing to a (possibly) new dimension to arrive at $\mathbf{x}''$, and then moving through another $k$ HMC steps with respect to the intermediate distribution $\pi_n^*$ in the new dimension to reach the final proposal $\mathbf{x}^*$. The forward transition can be written down as,
\begin{equation}
    \mathbf{x} \in \mathbb{R}^m \xrightarrow{\pi_m^*} \mathbf{x}'\in\mathbb{R}^m \xrightarrow{q_{mn}} \mathbf{x}'' \in \mathbb{R}^n \xrightarrow{\pi_n^*} \mathbf{x}^* \in \mathbb{R}^n
\end{equation}
where $q_{mn}$ denotes the proposal kernel to move from the initial dimension to the final dimension. Note that this could be stay, grow, or prune as defined previously. Similarly, the reverse move from $\mathbf{x}^*$ to $\mathbf{x}$ can be written as,
\begin{equation}
    \mathbf{x}^* \in \mathbb{R}^n \xrightarrow{\pi_n^*} \mathbf{x}''\in\mathbb{R}^n \xrightarrow{q_{nm}} \mathbf{x}' \in \mathbb{R}^m \xrightarrow{\pi_m^*} \mathbf{x} \in \mathbb{R}^m
\end{equation}
where $q_{nm}$ denotes the reverse proposal kernel.

Let $P_1$ and $P_2$ denote the transition kernels corresponding to the simulation of HMC through the intermediate distribution $\pi_m^*$ and $\pi_n^*$ respectively. Following the derivation presented in \citet{al2004improving}, to ensure detailed balance with respect to the overall  target distribution $\pi$, the following relationship must hold,
\begin{multline}
    \int_{(\mathbf{x},\mathbf{x}',\mathbf{x}'',\mathbf{x}^*) \in A \times B \times C \times D}  \pi_m(\mathbf{x}) P_1(\mathbf{x}'|\mathbf{x}) q_{mn}(\mathbf{x}''|\mathbf{x}') P_2(\mathbf{x}^*|\mathbf{x}'') \alpha_{mn} (\mathbf{x}^*|\mathbf{x}) d\mathbf{x} d\mathbf{x}' d\mathbf{x}'' d\mathbf{x}^*  \\
     = \int_{(\mathbf{x},\mathbf{x}',\mathbf{x}'',\mathbf{x}^*) \in A \times B \times C \times D} \pi_n(\mathbf{x}^*) P_2(\mathbf{x}''|\mathbf{x}^*) q_{nm}(\mathbf{x}'|\mathbf{x}'') P_1(\mathbf{x}|\mathbf{x}') \alpha_{nm} (\mathbf{x}|\mathbf{x}^*) d\mathbf{x} d\mathbf{x}' d\mathbf{x}'' d\mathbf{x}^*
\end{multline}
for Borel sets $A\in\mathbb{R}^m$, $B\in\mathbb{R}^m$, $C\in\mathbb{R}^n$, and $D\in\mathbb{R}^n$. This results in the following forward and backward acceptance probabilities,
\begin{equation}\label{eq:rjhmc-deriv-1}
    \alpha_{mn}(\mathbf{x}^*|\mathbf{x}) = \frac{\pi_n(\mathbf{x}^*)P_2(\mathbf{x}''|\mathbf{x}^*)q_{nm}(\mathbf{x}'|\mathbf{x}'')P_1(\mathbf{x}|\mathbf{x}')}{\pi_m(\mathbf{x})P_1(\mathbf{x}'|\mathbf{x})q_{mn}(\mathbf{x}''|\mathbf{x}')P_2(\mathbf{x}^*|\mathbf{x}'')}
\end{equation}

\begin{equation}\label{eq:rjhmc-deriv-2}
    \alpha_{nm}(\mathbf{x}|\mathbf{x}^*) = \frac{\pi_m(\mathbf{x})P_1(\mathbf{x}'|\mathbf{x})q_{mn}(\mathbf{x}''|\mathbf{x}')P_2(\mathbf{x}^*|\mathbf{x}'')}{\pi_n(\mathbf{x}^*)P_2(\mathbf{x}''|\mathbf{x}^*)q_{nm}(\mathbf{x}'|\mathbf{x}'')P_1(\mathbf{x}|\mathbf{x}')}
\end{equation}

Since $P_1$ satisfies detailed balance with respect to $\pi_m^*$, and similarly $P_2$ with respect to $\pi_n^*$, then the following relationships hold,
\begin{equation}
    \pi_m^*(\mathbf{x})P_1(\mathbf{x}'|\mathbf{x}) = \pi_m^*(\mathbf{x}')P_1(\mathbf{x}|\mathbf{x}') \quad\quad \pi_n^*(\mathbf{x}^*)P_2(\mathbf{x}''|\mathbf{x}^*) = \pi_n^*(\mathbf{x}'')P_2(\mathbf{x}^*|\mathbf{x}'')
\end{equation}

Substituting these expressions into Equations \ref{eq:rjhmc-deriv-1} and \ref{eq:rjhmc-deriv-2} results in the following expressions for the acceptance probabilities
\begin{equation}
    \alpha_{mn} (\mathbf{x}^*|\mathbf{x}) = \frac{\pi_n(\mathbf{x}^*)\pi_n^*(\mathbf{x}'')\pi_m^*(\mathbf{x})q_{nm}(\mathbf{x}'|\mathbf{x}'')}{\pi_m(\mathbf{x})\pi_n^*(\mathbf{x}^*)\pi_m^*(\mathbf{x}')q_{mn}(\mathbf{x}''|\mathbf{x}')},
\end{equation}

\begin{equation} 
    \alpha_{nm}(\mathbf{x}|\mathbf{x}^*) = \frac{\pi_m(\mathbf{x})\pi_m^*(\mathbf{x}')\pi_n^*(\mathbf{x}^*)q_{mn}(\mathbf{x}''|\mathbf{x}')}{\pi_n(\mathbf{x}^*)\pi_m^*(\mathbf{x})\pi_n^*(\mathbf{x}'')q_{nm}(\mathbf{x}'|\mathbf{x}'')}.
\end{equation}

Therefore, the acceptance probability for the double-sided transition becomes,
\begin{equation} \label{eq:app-rjhmc-tree-accept-prob}
    \alpha_{\textsc{accept,RJ}} = \min\left\{\frac{\pi_n(\mathbf{x}^*)\pi_n^*(\mathbf{x}'')\pi_m^*(\mathbf{x})q_{nm}(\mathbf{x}'|\mathbf{x}'')}{\pi_m(\mathbf{x})\pi_n^*(\mathbf{x}^*)\pi_m^*(\mathbf{x}')q_{mn}(\mathbf{x}''|\mathbf{x}')},1\right\},
\end{equation}
where $k$ HMC steps are taken on either side of the transition between dimensions. 

\newpage
\subsection{Hyperparameter Selection}\label{sec:app-hyp}

Table \ref{tab:grid-search} summarises the values investigated through the grid-search to determine the set of hyperparameters for each method tested on each relevant dataset.
Hyperparameters for CGM and WU were determined by taking the best-performing set of parameters with respect to the average testing accuracy (taken for a single chain after $N_{\textsc{warmup}}=500$ and $N=500$ samples) across the 5 folds. For those that had equal best accuracy values, the parameters that gave the best log-likelihood were selected. For SMC, hyperparameters were determined by taking the best-performing set with respect to the testing accuracy of the final set of tree particles (weighted appropriately) averaged across the 5 folds.

\begin{table}[H]
    \caption{Current set of hyperparameters run for the grid search. Parameters are labelled with respect to variable names in corresponding code implementations.}\label{tab:grid-search} 
\small\scshape
\begin{tabular}{lcm{10cm}}

    \toprule
    Method & Dataset & Hyperparameters \\
    \midrule
    \multirow{4}{*}{Wu} &  wisconsin & $\alpha = [2:1.0:10]$, $\beta = [1:1.0:12]$,  $\lambda=[0.6:0.2:1.2]$, $p=[0.3:0.2:0.7]$ \\
    & cgm & $\alpha = [2:1.0:12]$, $\beta = [8:1.0:12]$, $\mu_0 = [3:1.0:6]$, $n=[1:1.0:5]$, $\lambda = [0.6:0.2:1.2]$, $p = [0.3:0.2:0.7]$  \\
    & raisin & $\alpha = [1:1.0:6]$, $\beta = [1:1.0:5]$,  $\lambda=[0.6:0.2:1.8]$, $p=[0.3:0.2:0.7]$  \\
\midrule
\multirow{6}{*}{CGM}   &  iris & $\alpha = [0.1:0.1:0.5]$, $\alpha_\textsc{split} = [1.0:0.2:3.0]$, $\beta_\textsc{split} = [1.9:0.3:3.1]$ \\
        &  wisconsin & $\alpha = [1.0:0.5:2.5]$, $\alpha_\textsc{split} = [1.6:0.2:2.6]$, $\beta_\textsc{split} = [0.6:0.2:1.6]$ \\
        & cgm & $\alpha = [2:1.0:12]$, $\beta = [2:1.0:5]$, $\mu_0 = [3:1.0:6]$, $n=[0.1:0.1:0.5]$, $\alpha_\textsc{split} = [0.6:0.1:1.2]$, $\beta_\textsc{split} = [0.5:0.5:1.5]$  \\
        & wine &  $\alpha = [0.5:0.5:2.5]$, $\alpha_\textsc{split} = [0.6:0.2:1.0]$, $\beta_\textsc{split} = [0.8:0.2:1.6]$\\
        & raisin &  $\alpha = [0.5:0.5:2.5]$, $\alpha_\textsc{split} = [0.5:0.1:1.0]$, $\beta_\textsc{split} = [0.8:0.2:2.4]$\\
\midrule
\multirow{7}{*}{SMC}    & iris  & $\alpha_\textsc{split} = [0.5:0.1:1.0]$, $\beta_\textsc{split} = [0.5:0.1:1.5]$, $\alpha = [0.5:0.2:1.5]$, $n_\textsc{particles}=\{200,2000\}$, $n_\textsc{islands}=\{1,5\}$\\
        & wisconsin &  $\alpha_\textsc{split} = [0.6:0.2:1.6]$, $\beta_\textsc{split} = [0.1:0.2:1.0]$, $\alpha = [0.1:0.1:0.5]$, $n_\textsc{particles}=\{200,2000\}$, $n_\textsc{islands}=\{1,5\}$\\
        & wine &  $\alpha_\textsc{split} = [0.6:0.2:1.0]$, $\beta_\textsc{split} = [0.8:0.2:1.6]$, $\alpha = [0.1:0.1:0.5]$, $n_\textsc{particles}=\{200,2000\}$, $n_\textsc{islands}=\{1,5\}$\\
        & raisin &  $\alpha_\textsc{split} = [0.5:0.1:1.0]$, $\beta_\textsc{split} = [0.4:0.2:2.4]$, $\alpha = [0.5:0.5:2.5]$, $n_\textsc{particles}=\{200,2000\}$, $n_\textsc{islands}=\{1,5\}$\\
\bottomrule
\end{tabular}
\end{table}

\newpage
Table~\ref{tab:hyperparams} summarises the final set of hyperparameters used for different methods determined using the approach described above.

\begin{table}[h]
    \caption{Final hyperparameters used by methods for different datasets (options that are default values are not listed).}\label{tab:hyperparams}  
    \begin{center}
    \small\scshape
    \begin{tabular}{lcp{9cm}}
    \toprule
    Method & Dataset & Hyperparameters  \\
    \midrule
    \multirow{5}{*}{HMC-DF}    & iris & $h_\textsc{init}=0.01$, $h_\textsc{final}=0.01$,  $\alpha_\textsc{split} =0.45$, $\beta_\textsc{split} = 1.0$ \\
    & wisconsin & $h_\textsc{init}=0.025$, $h_\textsc{final}=0.025$,  $\alpha_\textsc{split} =0.45$, $\beta_\textsc{split} = 2.5$ \\
    & cgm & $h_\textsc{init}=0.001$, $h_\textsc{final}=0.001$,  $\alpha_\textsc{split} =0.45$, $\beta_\textsc{split} = 2.5$   \\
& wine & $h_\textsc{init}=0.025$, $h_\textsc{final}=0.025$,  $\alpha_\textsc{split} = 0.45$, $\beta_\textsc{split} = 2.0$  \\
& raisin & $h_\textsc{init}=0.005$, $h_\textsc{final}=0.001$,  $\alpha_\textsc{split} = 0.45$, $\beta_\textsc{split} = 2.0$  \\
\midrule
\multirow{5}{*}{HMC-DFI}    & iris & $h_\textsc{init}=0.01$, $h_\textsc{final}=0.01$,  $\alpha_\textsc{split} =0.7$, $\beta_\textsc{split} = 1.0$ \\
    & wisconsin & $h_\textsc{init}=0.1$, $h_\textsc{final}=0.025$,  $\alpha_\textsc{split} =0.95$, $\beta_\textsc{split} = 2.0$\\
    & cgm & $h_\textsc{init}=0.01$, $h_\textsc{final}=0.001$,  $\alpha_\textsc{split} =0.45$, $\beta_\textsc{split} = 2.5$   \\
& wine & $h_\textsc{init}=0.025$, $h_\textsc{final}=0.025$,  $\alpha_\textsc{split} =0.7$, $\beta_\textsc{split} =1.5 $  \\
& raisin & $h_\textsc{init}=0.05$, $h_\textsc{final}=0.001$,  $\alpha_\textsc{split} = 0.7$, $\beta_\textsc{split} = 2.5$  \\
\midrule
\multirow{3}{*}{WU} & wisconsin & $\alpha = 2.0$, $\beta = 2.0$, $\lambda = 0.8$, $p = 0.5$ \\
& cgm & $\alpha = 11.0$, $\beta = 11.0$, $\mu_0 = 3.0$, $n=1.0$, $\lambda = 0.8$, $p = 0.5$ \\
& raisin & $\alpha = 1.0$, $\beta = 2.0$, $\lambda = 1.8$, $p = 0.3$ \\
\midrule     
\multirow{5}{*}{CGM}   & iris &  $\alpha=0.1, \alpha_\textsc{split}=1.2, \beta_\textsc{split}=2.8$ \\
    & wisconsin & $\alpha=1.0, \alpha_\textsc{split}=2.6, \beta_\textsc{split}=1.2$\\
    & cgm & $\alpha = 4.0 $, $\beta = 2.0$, $\mu_0 = 3.0$, $n=0.3$, $\alpha_\textsc{split} = 1.2$, $\beta_\textsc{split} = 1.0$  \\
    & wine & $\alpha=1.0, \alpha_\textsc{split}=1.0, \beta_\textsc{split}=1.2$ \\
    & raisin & $\alpha=1.5, \alpha_\textsc{split}=0.7, \beta_\textsc{split}=2.2$ \\
\midrule     
\multirow{4}{*}{SMC}    & iris  & $\alpha_\textsc{split}=0.6, \beta_\textsc{split}=1.0, \alpha=0.7, n_\textsc{particles}=200, n_\textsc{islands}=1$ \\
    & wisconsin & $\alpha_\textsc{split}=1.6, \beta_\textsc{split}=0.7, \alpha=0.3, n_\textsc{particles}=200, n_\textsc{islands}=5$ \\
    & wine & $\alpha_\textsc{split}=1.0, \beta_\textsc{split}=1.0, \alpha=0.1, n_\textsc{particles}=200, n_\textsc{islands}=5$ \\
    & raisin & $\alpha_\textsc{split}=0.7, \beta_\textsc{split}=1.8, \alpha=0.5, n_\textsc{particles}=200, n_\textsc{islands}=5$ \\
    \bottomrule
\end{tabular}
\end{center}
\end{table}

\newpage

\end{document}